\documentclass[lettersize,journal]{IEEEtran}
\usepackage{amsmath,amsfonts}
\usepackage{algorithmic}
\usepackage{algorithm}
\usepackage{array}
\usepackage[caption=false,font=normalsize,labelfont=sf,textfont=sf]{subfig}
\usepackage{textcomp}
\usepackage{stfloats}
\usepackage{url}
\usepackage{verbatim}
\usepackage{graphicx}
\usepackage{cite}

\usepackage{booktabs} 
\usepackage{multirow}
\usepackage{color}
\def\red#1{\textcolor{black}{#1}}

\usepackage{amsmath,amssymb}

\begin{document}

\title{Unsupervised Visual Odometry and Action Integration for PointGoal Navigation in Indoor Environment}
\author{
    \IEEEauthorblockN{Yijun Cao,
    Xianshi~Zhang,
    Fuya~Luo,
    Chuan~Lin,
    and~Yongjie~Li,~\IEEEmembership{Senior~Member,~IEEE}}
\thanks{
Yijun Cao, Xianshi Zhang, Fuya Luo and Yongjie Li are with the MOE Key
Laboratory for Neuroinformation, the School of Life Science and Technology, University of
Electronic Science and Technology of China, Chengdu 610054, China.
E-mail: yijuncaoo@gmail.com, zhangxianshi@uestc.edu.cn, luofuya1993@gmail.com,
liyj@uestc.edu.cn.(\textit{Corresponding authors: Xianshi Zhang.})}
\thanks{Chuan Lin is with the College of Electric and Information
Engineering, Guangxi University of Science and Technology, Liuzhou 545006,
China. E-mail: chuanlin@gxust.edu.cn.}
\thanks{This work was supported by STI2030-Major Projects (2022ZD0204600), Sichuan Science and Technology Program (2022ZYD0112)
and National Natural Science Foundation of China (62076055).}}

\markboth{Journal of \LaTeX\ Class Files,~Vol.~14, No.~8, August~2021}%
{Shell \MakeLowercase{\textit{et al.}}: A Sample Article Using IEEEtran.cls for IEEE Journals}


\maketitle

\begin{abstract}
    PointGoal navigation in indoor environment is a fundamental task for personal robots to navigate to a specified point.
    Recent studies solved this PointGoal navigation task with near-perfect success rate in photo-realistically simulated environments,
    under the assumptions with noiseless actuation and most importantly, perfect localization with GPS and compass sensors.
    However, accurate GPS signalis difficult to be obtained in real indoor environment.
    To improve the PointGoal navigation accuracy without GPS signal,
    we use visual odometry (VO) and propose a novel action integration module (AIM) trained in unsupervised manner. 
    Sepecifically, unsupervised VO computes the relative pose of the agent from the re-projection error of two adjacent frames, 
    and then replaces the accurate GPS signal with the path integration.
    The pseudo position estimated by VO is used to train action integration 
    which assists agent to update their internal perception of location and helps improve the success rate of navigation.
    The training and inference process only use RGB, depth, collision as well as self-action information.
    The experiments show that the proposed system achieves satisfactory results and
    outperforms the partially supervised learning algorithms on the popular Gibson dataset.
\end{abstract}


\section{Introduction}\label{sec_introduction}
Considering how a robot placed in a novel indoor environment can navigate to a target point, 
e.g., ``goes 2 meters north, 5 meters west relative to the start".
This task, known as PointGoal navigation, 
requires the agent to search through the environment unvisited before and approach the target point.

As a basic navigation task in indoor environment, 
recent methods solve it with near-perfect accuracy (99.6\% success) \cite{wijmans2019dd}
under the assumptions of noiseless egocentric action and accurate localization using GPS and compass sensors.
However, these assumptions are difficult to hold in a conventional indoor navigation environment,
because the motion process of agents involves physical (e.g., motors and gears), and environmental errors (e.g., collision),
which may introduce motion uncertainty to the navigation algorithm.
In addition, GPS sensors typically yield an unsatisfactory localization accuracy in indoor environments.
Considering these realistic settings, recent researches try to use visual odometry (VO) \cite{zhao2021surprising} to replace the GPS sensors
or hybrid simultaneous localization and mapping (SLAM) with planning approach \cite{karkus2021differentiable} 
for building embedded navigation system. 
However, these methods still use accurate GPS information as supervision to train their VO or SLAM systems. 
Thus, in practical indoor applications, it is a promising research direction to train a visual navigation model that completely discards GPS signals.

Navigation only using self-perception is an innate ability of many animals.
They are capable of navigating in complex environments, finding food and going back to their nests \cite{mandal2018animals}.
Many researches have shown that some neurons in the brain are closely related to the animal's ability to navigate, 
for example the place cells \cite{o1978hippocampal}, head direction cells \cite{taube1990head} and grid cells \cite{hafting2005microstructure}.
These cells guide their navigational activities by producing specific responses based on their specific position and orientation in space.
The navigational activities also rely on two fundamental mechanisms: path integration and landmark calibration \cite{etienne1996path}.
Through path integration, animals update their internal neural representations of place using self-motion information.
However, the path integration using motion information alone may lead to rapid cumulative errors in both the direction and
distance to the goal. 
Thus, landmark calibration mechanism is further employed to help the animals correct their cumulative errors caused by the inaccuracy path integration \cite{etienne1996path}. 
Inspired by these biological findings, 
we use VO and motion integration to estimate the motion, and expect to replace inaccurate GPS signal in indoor environments.

Training an unsupervised VO model can be implemented by 
computing re-projection error from two consecutive frames \cite{zhou2017unsupervised,wang20223d}.
In practice, an unsupervised method will encounter two key problems compared with supervised manners.
The first one is the scale uncertainty. 
Monocular RGB images cannot provide the specific scale of motion, e.g., centimeter or meter, so we need depth sensor to provide an absolute scale. 
Fortunately, depth sensor in indoor environment can provide quite accurate depth information, 
so we also use the data of depth to train the model.
The second is the accuracy of monocular visual odometry.
Compared with supervised methods, the unsupervised algorithms rely 
only on re-projection error and are therefore difficult to converge to the optimal value. 
To address this issue, we 
1) use richer information to construct re-projection error;
2) propose a novel classified Pose-Net (CP-Net) to estimate ego-motion with Linear Motion (LM) probability volumes 
which transforms the regression prediction into a probability sum of classification prediction.

The action integration model (AIM) estimates ego-motion at each moment by its own action.
Compared with visual model, the uncertainty of the action integration is greater, 
thus we use the pose estimates obtained from VO as pseudo-labels to train the action model.
This strategy is based on two key ideas: 
1) using LSTM \cite{hochreiter1997long} network to compute action-only path integration, which receives the previous action and collision information, and 
2) using accumulative VO as pseudo location label to supervise action path integration.
AIM can not get an accurate results due to 
inputting a noisy action and using inaccurate pseudo label as supervision.
Thus, we hope that this component can help VO by acting colibration as auxiliary positioning or
teaching the intelligent agent to be able to think more about exploration in its next decision, 
rather than being driven only by the target location.
Motivated by the grid cell representation, 
which proves helpful for PointGoal navigation by making the agent to learn to take shortcuts in some situations \cite{banino2018},
we use place and head direction cells found in both mouse and human brains \cite{moser2014grid} to encode the pseudo location label.

To the best of our knowledge, 
this work is the first method that does not use position signals (GPS+Compass), 
neither for training nor for inferring in the learning based PointGoal navigation task.
The main contributions are as follows:
\begin{itemize}
    \item 1) An unsupervised VO method is designed to train a VO model using re-projection error.
    The VO model is improved by using the proposed CP-Net and richer information to estimate ego-motion.
    \item 2) A new AIM model is built based on the LSTM network 
    and trained using the predicted VO as pseudo ground truth label for unsupervised training.
    \item 3) Experimental results show that the features outputed by the AIM model 
    can significantly improve the navigation performance.
\end{itemize}

On the Habitat simulation platform \cite{savva2019habitat} with the widely used Gibson real-world indoor scene dataset \cite{xia2018gibson},
our experiments demonstrate that the proposed unsupervised learning algorithm is feasible in PointGoal navigation task 
and outperforms those partially supervised learning algorithms in quantitative and qualitative comparisons.

\begin{figure*}[t]
    \centering
    \includegraphics[width=17cm]{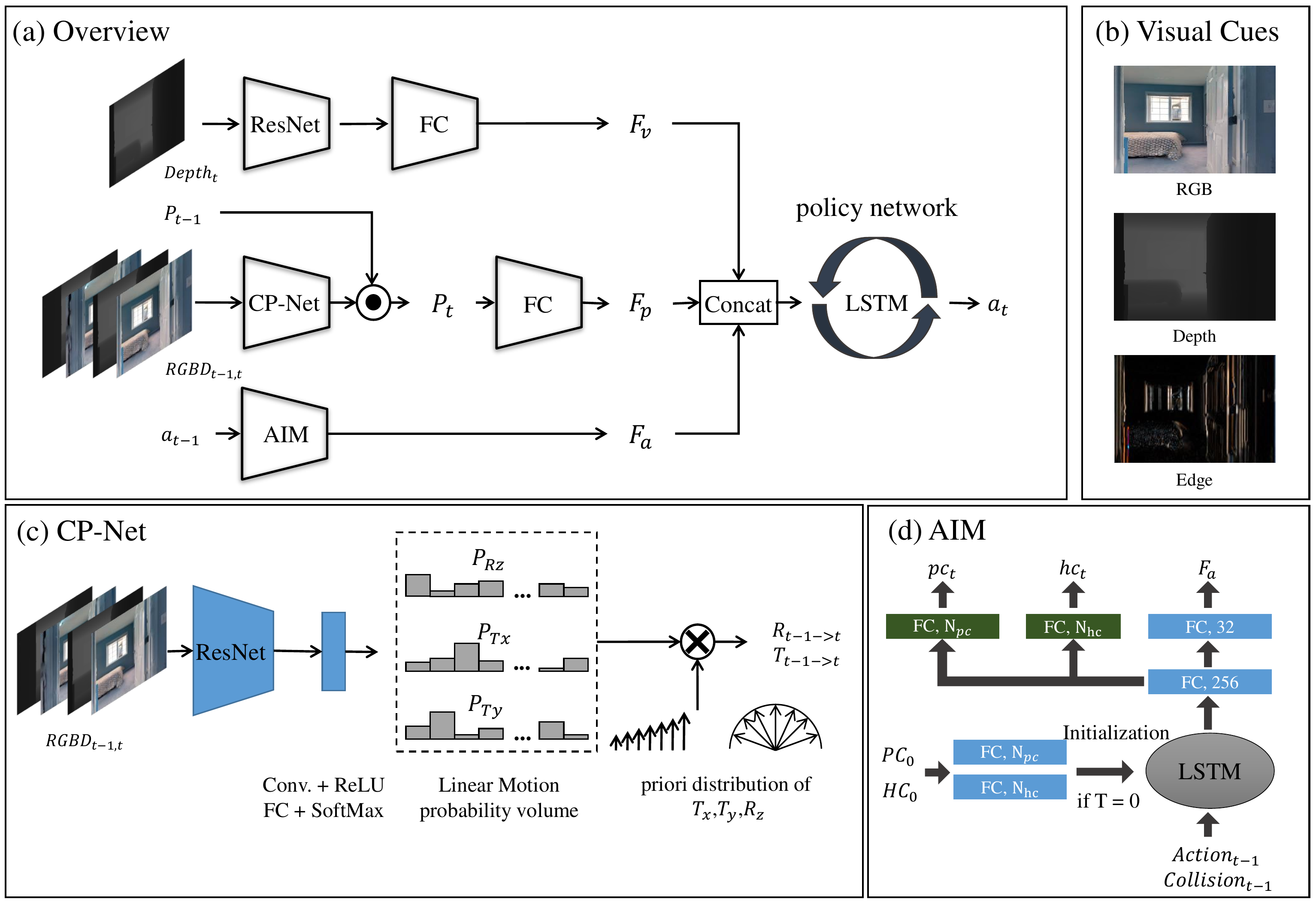}
    \caption{(a) The policy network is a 2-layer LSTM, 
    which receives observational visual feature ($F_v$),
    current position feature ($F_p$), and previous action feature ($F_a$).
    $F_v$ is encoded by ResNet18 \cite{he2016deep};
    $F_p$ provides position information which is predicted by the proposed CP-Net with visual path integration;
    $F_a$ provides action and self-motion information which is predicted by the proposed AIM with self-motion path integration.
    (b) Visual cues. We use RGB, depth and vertical edge of RGB image as the visual cues for training our unsupervised VO model.
    (c) The proposed CP-Net. The CP-Net is composed of a ResNet18 \cite{he2016deep} 
    and a classification head activated by SoftMax function,
    mapping the input two frames to (LM) probability volume $f: RGBD_{t-1, t} \rightarrow \{R_{pz}, T_{px}, T_{py}\}$.
    Then, the probability volume is converted to a rotation $R_{t-1 \rightarrow t}$ and 
    a translation $T_{t-1 \rightarrow t}$ in 2D plane.
    (d) The proposed action integration model (AIM).
    The model inputs are previous action and collision, the outputs are $pc_t$, $hc_t$ and action feature $F_a$
    where $pc_t$ and $hc_t$ are the predicted place and head-direction cell representations, respectively.
    The subscript $t$ indicates the $t$-$th$ moment of recurrent LSTM.
    $PC_0$ and $HC_0$ are initialized representations of place, and head-direction cell, respectively.
    }
    \label{fig:architecture}
\end{figure*}

\section{Related works}\label{sec_related_works}
\textbf{Learning-based navigation.}
Conventional navigation methods usually decompose the problem into two separate stages:
building a metric or topological map of the environments, and planning a path to the goal \cite{thrun1998learning}.
Benefitting from the powerful ability of deep neural networks, many researches turn to reinforcement learning (RL) \cite{wijmans2019dd, ding2022monocular}
or learning based SLAM \cite{chaplot2019learning,karkus2021differentiable} to help agents make action decisions. 

Recently, there has been a renewed interest in the field of embodied navigation tasks in indoor environment,
which can be distinguished into PointGoal navigation \cite{wijmans2019dd,zhao2021surprising}, 
ImageGoal navigate \cite{kwon2021visual}, ObjectGoal navigation \cite{du2020vtnet,zhang2021hierarchical} 
and language-guided navigation \cite{zhang2020language}.
For the PointGoal navigation task, the agent is asked to travel to a given coordinate point.
To solve this problem, some researchers construct a grid map with SLAM technology, 
and use path planning algorithms to find local targets, thereby continuously approximating the global target \cite{chaplot2019learning,karkus2021differentiable};
others estimate action using RL with visual features and predicted odometry information \cite{zhao2021surprising,gupta2017cognitive}.
ImageGoal and ObjectGoal tasks require intelligent agents to respectively find a specific image and target object, which
proposes a greater challenge to explore in unfamiliar environments.
Currently, researchers have tried to use dynamical topological maps \cite{kwon2021visual,zhang2021hierarchical},
spatial attention mechanism \cite{mayo2021visual}, visual transformer \cite{du2020vtnet}, 
or introduce auxiliary exploration targets \cite{ye2021auxiliary,maksymets2021thda} to address such two tasks.
Language-guided navigation, or named vision-and-language (VLN), uses linguistic goals instead of metric goals to do the navigation task.
Zhang et al. \cite{zhang2020language} designed a cross-modal grounding module and proposed to recursively alternate
the learning schemes of imitation and exploration.
One of the biggest challenges for VLN is to associate natural language with visual input while keeping track of which part of the instruction has been completed.
To address these issues, many methods rely on visual-textual alignment \cite{deng2020evolving}, 
attention mechanisms \cite{chen2021topological} or adapt powerful language models \cite{georgakis2022cross} to the VLN task. 

Although these tasks differ in their setups, 
each of them requires the agent to navigate accurately in the environment. 
For this reason, the agent's navigation strategy assumes perfect localization of the agent's position and orientation
(e.g., a perfect GPS+Compass sensor). 
To alleviate this unrealistic assumption, 
Zhao et al \cite{zhao2021surprising} proposed to estimate self-motion from a pair of RGBD maps with supervised training. 
Our work is similar to \cite{zhao2021surprising}. 
But quite different from that work, we use unsupervised strategy to train the VO and AIM systems.

\textbf{Visual odometry.} Visual odometry (VO) is a long-standing problem that estimates the ego-motion incrementally using visual input. 
A classical geometry-based VO system usually consists of two steps. 
First, the raw camera measurements are processed to generate a photometric \cite{Engel-pami18} or feature \cite{Mur-tro17} representation. 
Second, the representation is used to estimate depth and ego-motion using geometry methods (e.g., epipolar geometry and triangulation \cite{Mur-tro17}) 
and local optimization (e.g., photometric BA \cite{Engel-pami18}). 
Recently, many researchers have tried to solve the VO problem using convolutional neural networks (CNNs) in a supervised or unsupervised manner \cite{Godard-iccv19}. 
The supervised methods minimize the distance between predicted values (depth and ego-motion) and corresponding ground truth by using 
such strategies as a recurrent neural network (RNN) \cite{Wang-icra17}, memory mechanism \cite{Xue-cvpr19}, or feature-metric BA \cite{Tang-iclr19}.

In contrast, to avoid the need for annotated data, unsupervised VO has been developed using the standard structure-from-motion pipeline. 
These methods accept continuous image input and infer VO via a CNN \cite{Zhou-cvpr17} or LSTM \cite{Zou-eccv20}.
To combine the advantages of geometry-based and deep-learning methods, several works \cite{Zhan-icra19,cao2023learning} 
tried to learn the various components (e.g., optical flow, depth, and VO) of the entire system to get more accurate performance.
To further improve the utilization of training data, 
Wang et al. \cite{wang20223d} proposed a 3D hierarchical refinement and augmentation method.
Chen et al. \cite{9386100} exploited the point cloud consistency constraint 
and use threshold masks to filter dynamic and occluded points aiming to overcome the effect of light transformation on performance.
Wei et al. \cite{wei2021iterative} proposed an iterative feature matching framework
to avoid local minima in non-texture or repeated-texture environments.
Some works tried to improve the network structure. 
For example, Tian et al. \cite{tian2021depth} proposed a novel network using quadtree constraint
and Song et al. \cite{song2021monocular} used Laplacian pyramid into the decoder network.

Similar to \cite{Godard-iccv19, 9386100, tian2021depth, song2021monocular}, 
which use standard structure-from-motion pipeline jointly training depth and VO,
we further improve the VO performance with richer visual cues and the proposed CP-Net architecture to training more accurate VO.

\section{Method}\label{sec_method}
\subsection{Overview} \label{subsec_overview}
The overall architecture of our model is illustrated in Figure \ref{fig:architecture}(a).
Given a set of environments $E$ and point goals $G$, in each navigation episode, 
the agent is initialized at random location $\{\rho, \theta\}$ in an environment $e \in E$,
where $\rho$ and $\theta$ represent respectively the distance and yaw angle related to the goal. 
At each time step $t$, the agent predicts an action $a_t \in A$ via a policy network $\pi (F_v, F_p, F_a)$,
where $F_v$ is visual feature which receives current depth map and is encoded by a fully connected layer after ResNet18 \cite{he2016deep};
$F_p$ is position feature which receives the current position $p_t$ predicted by the proposed VO model and encoded by a fully connected layer;
$F_a$ is action feature which receives previous action $a_{t-1}$, previous collision signal $co_{t-1}$ and
is encoded by the proposed motion path integration model.
The agent computes a distribution over an action space $A = \{\textit{move forward}, \textit{turn left}, \textit{turn right}, \textit{stop}\}$,
and the translation and rotation steps are set to $0.25m$ and $10^{\circ}$, respectively.
The success of object navigation task requires the agent finally to get close to the point goal (less than a threshold).

\subsection{Visual feature extraction}\label{subsec_fv}
We use ResNet-18 \cite{he2016deep} as our visual feature extractor to process an egocentric observation with a size of
341(width) $\times$ 192(height). Following \cite{wijmans2019dd,zhao2021surprising}, 
we replace every BatchNorm \cite{ioffe2015batch} layer with GroupNorm \cite{wu2018group} to deal
with highly-correlated trajectories in on-policy RL and massively distributed training. A 2x2-AvgPool layer is added
before ResNet-18 so that the effective resolution is $170 \times 96$.
ResNet-18 produces a $256 \times 6 \times 3$ feature map, which is
converted to a $114 \times 6 \times 3$ feature map through a 3x3-Conv layer.
After ResNet-18 encoder, $F_v$ is obtained by a fully connected (FC) layer with ReLU as activation function.

\subsection{Unsupervised visual odometry}\label{subsec_unvo}
At each time step $t$, the VO model estimats 
the rotation $R_{t-1 \rightarrow t}$ and translation $T_{t-1 \rightarrow t}$ between $t-1$ and $t$ in a 2-dimensional (2D) plane.
Given the previous position ($P_{t-1} \in \mathcal{R}^2$), current position $P_{t}$ can be formulated as:
\begin{equation}
    {P}_{t} = R_{t-1 \rightarrow t}{P}_{t-1} + T_{t-1 \rightarrow t}.
\end{equation}

After obtaining the position ${P_t}$ related to the start point, 
we first transform it to the point related to the target goal $P^{gps}_t=P^{gps}_0 - P_t$,
and then convert the point representing as cartesian coordinates 
to the distance $\theta_t$ and direction $\rho_t$ using polar coordinates.
The final position feature $F_p$ is obtained by a FC layer (32-dimensional feature) 
with $\theta_t$, $cos(\rho_t)$ and $sin(\rho_t)$ as inputs.

\textbf{Unsupervid learning pipeline.}
The core of unsupervised training pipeline is to find the corresponding pixels with regard to depth.
Given a pair of adjacent observations $O_{t}$ and $O_{t'}$,
for a coordinate of pixel $c_{t}$ in $O_{t}$, 
the corresponding pixel $c_{t'}$ in $O_{t'}$ can be found 
through camera perspective projection for static scenes.
Formally, the relationship can be written as
\begin{equation}
    c_{t'} = K(R_{t \rightarrow t'}D_{t}(c_{t})K^{-1}c_{t} + T_{t \rightarrow t'}), \\
\end{equation}
where $K$ is the camera intrinsic, and $D_{t}(c_{t})$ denotes the depth at the coordinate $c_{t}$.
$R_{t \rightarrow t'}$ and $T_{t \rightarrow t'}$ are respectively 
the rotation and translation of camera pose from time stamp $t$ to $t'$.

After obtaining the corresponding $c_{t}$ and $c_{t'}$, 
the observation $O_{t'\rightarrow t}$ can be synthesized using $I_{t'}$
by warping the target coordinate $c_{t'}$ into the source $c_{t}$.
Then, unsupervised training strategy is realized by minimizing the
photometric re-projection error between $O_{t}$ and the synthetic observation $O_{t' \rightarrow t}$:
\begin{equation}
   \mathcal{L}_{self} = \frac{1}{|V|}\sum_{c_t \in V} \min_{t'} r(O_t(c_t), O_{t' \rightarrow t}(c_t)),
   \label{eq:loss_vo}
\end{equation}
where $V$ is the available pixels for computing the error. 
The function $r(O_t(c_t), O_{t' \rightarrow t}(c_t))$ is the metric
between the source observation $O_t(c_t)$ and the synthetic observation $O_{t' \rightarrow t}(c_t))$.
Similar to \cite{godard2019digging}, 
the metric is defined as mean absolute error (MAE) with structural similarity SSIM \cite{Zhou-tip04};
$O_{t'}$ contains two adjacent temporal frames around $O_t$, i.e., $I_{t'} \in \{ I_{t-1}, I_{t+1} \}$.

\textbf{Training with richer visual cues.}
Conventional unsupervised learning methods \cite{cao2023learning,godard2019digging} 
use RGB image $O=[R,G,B]_t \in \mathcal{R}^{H \times W \times 3}$ as the synthesized target.
For estimating more accurate pose, we use richer visual cues as metrics between synthetic and original targets.
The first cue is depth. The advantage is that depth information 
is not easily disturbed by luminance and complex textures.
The second cue is structured texture. 
Considering that the robots can only move on the ground, 
the edge information in the vertical orientation has the greatest visual discrimination and 
therefore also facilitates the distinction between the synthetic and the original targets.
Inspired by the primary visual cortex of biological visual system \cite{hubel1962receptive}
which can fast extract multi-scale and mult-orientation edges \cite{zeng2011center},
we use the bio-inspired contour detection model \cite{cao2019application} to detect vertical edges
as a kind of auxiliary visual cue. 
In summary, the synthesized target consists of RGB image, depth map and vertical edges of RGB image, 
as shown in Figure \ref{fig:architecture}(b):
\begin{equation}
    \begin{split}
        O&= [RGB,depth,edge] \in \mathcal{R}^{H \times W \times 7} , \\
        edge &= RGB(x,y) \ast \frac{\partial g(x,y,\theta_g,\sigma)}{\partial x},
    \end{split}
    \label{eq:visual_cues}
\end{equation}
where the symbol $\ast$ denotes the convolution operation; $\theta_g=\pi/2$ and $\sigma=1.0$ are set for extracting vertical edge with fine scale.
In addition, our experiment has shown that SSIM is not suitable for measuring the differences in depth and edges of local regions, 
because the local averaging of this operation leads to less differentiation in local images.
So the total metric function in this work is defined as:
\begin{equation}
    \resizebox{.91\linewidth}{!}{$
    \displaystyle
    \begin{aligned}
        r(\Delta  O) = \sum_{C \in O} 
            \begin{cases} 
                \alpha MAE(\Delta C) + \beta SSIM(\Delta C) &C = RGB; \\  
                \gamma MAE(\Delta C) &otherwise,
            \end{cases}
    \end{aligned}
    $}
    \label{eq:visual_loss}
\end{equation}
where $\alpha=0.15$, $\beta = 0.85$, $\gamma=1.0$, $\Delta O$ and $\Delta C$ are abbreviations for 
$(O_t, O_{t' \rightarrow t})$ and $(C_t, C_{t' \rightarrow t})$, respectively.

\begin{figure*}[t]
    \centering
    \includegraphics[width=16cm]{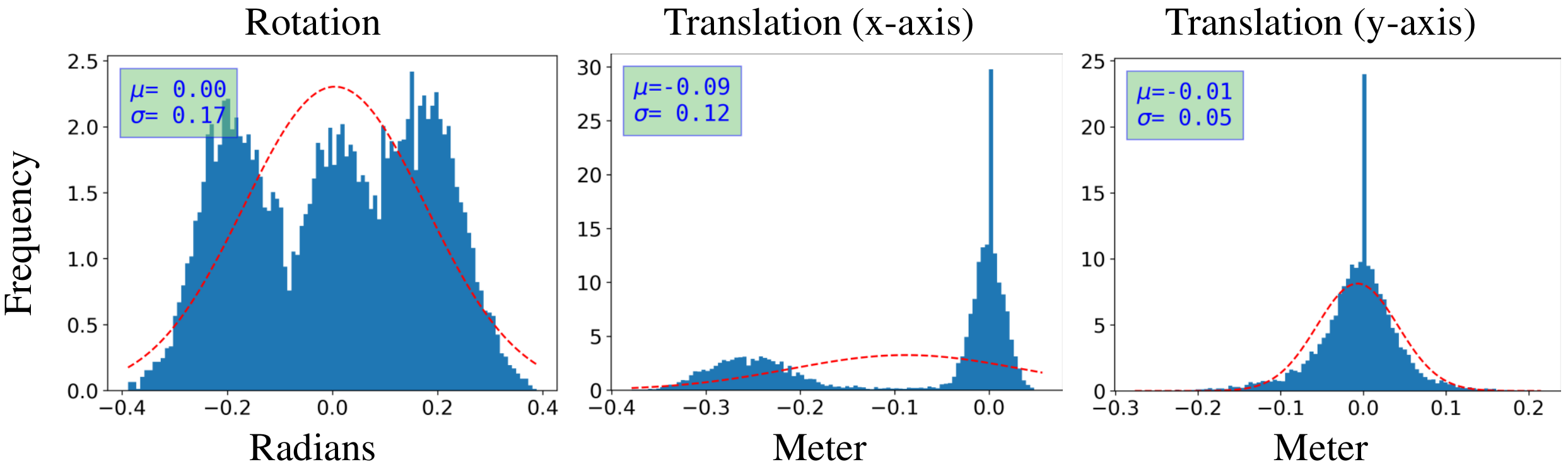}
    \caption{Translation and rotation distribution histogram of each action in our VO training dataset. 
    }
    \label{fig:action_statics}
\end{figure*}

\textbf{Network.}
The proposed CP-Net, as shown in Figure \ref{fig:architecture}(c), 
is composed of a ResNet18 \cite{he2016deep}, 3 convolutional layers with ReLU function, and 3 FC layers with \textit{SoftMax} function.
The first layer of the ResNet18 is modified as 8 channels in order to take two RGBD frames as inputs.

Each frame includes bimodality cues, i.e., RGB image and depth map. 
As visual information, both of them can be extracted and fused by convolutional neural networks. 
In order to ensure and promote better fusion of meaningful message, 
different tasks usually employ different approaches, which generally include input fusion, feature fusion and output fusion.
Chen et al. \cite{chen2018progressively} analyzed the role of input fusion and 
feature fusion in salientobject detection task with RGBD inputs, 
and found that input fusion for RGBD cues is better than single RGB image, and feature fusion method is better than input fusion.
Cao \cite{caolearning} used output fusion approach for contour detection task, 
indicating that the result is still better than that by the model with a single RGB message. 
Although fusion at the input layer is usually less effective than feature fusion method, 
this paper verifies that there is still some improvement over unimodal information.

The network maps the two input frames of RGBD images to three grouped vectors, 
$f: RGBD_{t-1, t} \rightarrow \{P_{Rz}, P_{Tx}, P_{Ty}\}$.
The three grouped vectors $P_{Rz}, P_{Tx}, P_{Ty}$ are Linear Motion (LM) probability volumes,
and each element of the volume represents the probability of the priori distribution. 
When given the priori value of translation along x-axis $T_x$, y-axis $T_y$, and rotation $R_z$,
the final predicted translation $T_{t-1 \rightarrow t}$ and rotation $R_{t-1 \rightarrow t}$ 
are obtained by computing the expectation, 
\begin{equation}
    T_{t-1 \rightarrow t} = 
    \begin{bmatrix}
        T_x {P_{Tx}}^\intercal & T_y {P_{Ty}}^\intercal
    \end{bmatrix} ^{\intercal},
\end{equation}

\begin{equation}
    R_{t-1 \rightarrow t} = 
    \begin{bmatrix}
        cos(R_z {P_{Rz}}^\intercal) & -sin(R_z {P_{Rz}}^\intercal) \\
        sin(R_z {P_{Rz}}^\intercal) & cos(R_z {P_{Rz}}^\intercal)
    \end{bmatrix},
\end{equation}
where $R_{t-1 \rightarrow t} \in \mathbb{S} \mathbb{O}(2)$ denotes the estimated rotation matrix from the 2D special
orthogonal group;
$T_x$, $T_y$, and $R_z$ are set to uniformly distributed, which are generated using arithmetic sequences 
$[start + (i-1) * step]$, $i=1,...,N$.
According to the statistics of the VO dataset (Figure \ref{fig:action_statics}), 
we found that the robot's motion is within a fixed range when it is given a certain motion command.
Thus, the parameters $[start, N, step]$ of $T_x$, $T_y$, and $R_z$ are set to $(-0.4, 81, 0.01)$, 
$(-0.25, 51, 0.01)$, 
$(-0.1, 51, 0.01)$, respectively.
The LM probability volume can be understood as a way of pose discretization. 
Previous works \cite{godard2019digging,zhou2017unsupervised} generally regard the pose estimation as a regression task, 
which uses a CNN to predict the orientations and translations.
In contrast, based on the statistics of agent's movements, 
the proposed CP-Net splits all possible movement amplitudes to small bins 
and then predicts their probabilities.
According to our experiments, this operation helps obtain better numerical accuracy than regression models, 
which may be due that the refined small bins have greater discriminatory power around the optimal value.

In this paper, we directly use the plain concat operation to fuse RGB and depth as inputs to the VO network. 
This solution is the simplest but not necessarily the best one. 
Despite the simplicity of the method, direct concat can still achieve a good result (see Table II) 
because both RGB and depth are visual information that can be normalized to 0-1 without destroying any information, 
and their local relationships can be captured by the convolutional neural network to make predictions.


\subsection{Action integration}\label{subsec_motion_model}
Existing works \cite{zhao2021surprising, wijmans2019dd} only use previous action as currently moving cues to train the policy network.
Inspired by mammal's spatial localization mechanisms \cite{hafting2005microstructure, etienne1996path}, 
we argue that a better location requires not only the visual landmark of the current moment, 
but also the inertial motion state of the previous moment. 
Also inspired by \cite{banino2018}, 
which uses translational and angular velocities as inputs and trains a grid cell representation using LSTM with supervision,
and this representation endows agents with the ability to perform vector-based navigation 
and improves the convergence speed of policy network.
Thus, we use action integration to indicate the inertial motion of the agent and use LSTM as action integration model (AIM).

In this work, we propose the AIM that
uses pseudo trajectory obtained by vision to correct action integration.
Specifically, we use LSTM, as shown in Figure \ref{fig:architecture}(d),
which is required to update its estimate by processing 
discrete action and collision signals at previous time step $t-1$.
AIM is not only supervised by pseudo position label (predicted by VO model), but also updated by task driven policy gradient.

\textbf{Nerual Representations.}
Place and direction are typically represented respectively as Cartesian coordinates and angles.
However, such representations are not consistent with place and head-direction cells in mammals \cite{taube1998,moser2008}.
Inspired by \cite{banino2018}, we use a neurocompatible encoding method to encode the raw format as a neural vector representation.

Place cell activations $PC \in [0, 1]^{N_{pc}}$, encoded from Cartesian coordinates $\mathcal{X} = (x, y)$, 
are simulated by a 2D Gaussian function with standard deviation $\sigma$ activated by SoftMax function, written as
\begin{equation}
    PC = softmax(-(\mathcal{X} - \overrightarrow{\mu_p})^2 / 2\sigma^2 )
\end{equation}
where $\overrightarrow{\mu_p} \in \mathbb{R}^{N_{pc}}$ are $N_{pc}$ 2D vectors chosen uniformly before training, 
and $\sigma=0.5$, the place cell scale, is a positive scalar fixed before training. 
For ensuring that this representation can cover the whole environment, 
the places are uniformly distributed throughout the environment.
In our experiments, the range of $(x,y)$ in Gibson dataset is $[-25m, 25m]$, 
and thus the number $N_{pc}$ is determined by $\sigma$: $N_{pc}=((25*2)/5\sigma)^2$.

For a given angle $\theta$, head-direction cell activations $HC \in [0, 1]^{N_{hc}}$ 
are represented by a Von Mises distribution activated by SoftMax function,
\begin{equation}
    HC = softmax(e^{k \cos \pi (\theta - \overrightarrow{\mu_h}) / 180}),
\end{equation}
where $k=20$, $N_{hc}=12$, direction centers $\overrightarrow{\mu_h} \in [0, 360)$ are sampled uniformly before training.

\textbf{Network.}
As shown in Figure \ref{fig:architecture}(d), the inputs of the proposed model are discrete action $a_{t-1}$ and 
collision $co_{t-1}$, which are respectively encoded as a vector using embedding layer.
The hidden states of the LSTM, $l_0$ and $m_0$, are initialized respectively by computing a fully connected layer
of the ground truth neural vector representation of place $PC_0$ and head-direction $HC_0$ cells at time 0.
Note that the position is initialized by the GPS signal at the first time step, 
and the direction is initialized as zero.

The LSTM outputs are sent to a fully connected layer with 256-D outputs.
The outputs of the proposed model consist of three branches, 
one of them computes the action feature $F_a$ as the input of policy network.
The other two predict respectively the place and head-direction cell activations, 
activated by SoftMax function with $N_{pc}$-D and $N_{hc}$-D outputs.

\textbf{Loss.}
Given the pseudo position $PC$ encoded using vector representation, the AIM is trained under the constraint
\begin{equation}
    \mathcal{L}_{vc} = \omega_{pc} \mathcal{L}_{pc} + \omega_{hc}\mathcal{L}_{hc},
    \label{AIM_loss}
\end{equation}
where $\mathcal{L}_{pc}$ is the cross-entropy loss between pseudo position $PC_t$ and AIM's output $pc_t$ at time step $t$;
$\mathcal{L}_{hc}$ is the cross-entropy loss between pseudo direction $HC_t$ and AIM's output $hc_t$ at time step $t$.
Both $\omega_{pc}$ and $\omega_{hc}$ are set to 0.05.

\subsection{Navigation policy} 
The navigation policy (Figure \ref{fig:architecture}(a)) consists of a 2-layer LSTM \cite{hochreiter1997long},
which receives observational visual feature $F_v$, action feature $F_a$ outputted by the proposed AIM, 
and position feature $F_p$ predicted by the proposed VO model.
At each time step $t$, the policy $\pi(\cdot)$ operates on these features and computes a distribution over the action space $A$. 
To learn the policy, we use the DD-PPO algorithm \cite{wijmans2019dd} with the same set of hyper-parameters and reward shaping settings.

\section{Experiments}\label{sec_exp}
\subsection{Experimental setup}
\textbf{Simulator and Datasets.} Our experiments were conducted on the Habitat simulator \cite{savva2019habitat}, 
which is a 3D simulation platform for embodied AI research.
Based on Habitat simulator, we use the Gibson dataset \cite{xia2018gibson}, which is widely used in the field of vision navigation. 
Savva et al. \cite{savva2019habitat} evaluated the Gibson dataset by rating every mesh reconstruction on
a quality scale of 0 to 5 and then collected all splits such that each only contains scenes with a rating of
4 or above (Gibson-4+). 
The Gibson-4+ dataset has total 86 scenes, all models were trained with 72 of these scenes, 
and evaluated on 14 unseen scenes according to the split used in \cite{savva2019habitat}.

\textbf{VO dataset construction.}
To train the VO model, we created a dataset of 100,000 steps from 13,036 trajectories randomly sampled from 72 training scenes.
Each step consists of an observation with RGBD sensor.
The data collection process includes the following two steps: 
1) randomly sampling a starting position and orientation of the agent and a navigable PointGoal in the scene; 
2) collecting the shortest path to navigate from the starting point to the point goal.
After the VO model is trained, we fixate the VO module and fine-tune the policy network with vision-motion calibration loss 
by using the pre-trained model from \cite{wijmans2019dd}.

\textbf{Agent.}
The agent was equipped with an RGBD camera mounted at a height of $0.88m$. 
It has a $70 ^{\circ}$ field of view and 
records egocentric observations with a resolution of $192 \times 341$ pixels for policy network and $256 \times 256$ pixels for VO model.
The action space $A$ consists of four actions: \textit{move forward} ($\backsim 0.25m$),
\textit{turn left} and \textit{turn right} ($\backsim 10^\circ$), and \textit{stop}. 
The agent exhibits actuation action noise modeled by LoCoBot robot \cite{murali2019pyrobot}. 
During collisions, the "sliding" behavior that allows the agent to slide along the obstacle instead of stopping is disabled. 

\textbf{Training.}
Both the policy network and CP-Net of the proposed work are implemented with PyTorch \cite{paszke2017automatic} 
and trained on a single RTX 3090 GPU.
Training setting of the policy network is the same as \cite{wijmans2019dd}. 
The VO model uses Adam \cite{kingma2014adam} optimizer with $\beta_1=0.9$ and $\beta_2=0.999$.
The learning rate of the CP-Net is $10^{-4}$ and the batch size is set to 32.

The entire training schedule consists of two stages:
(1) stage I: self-supervised training of visual odometry model using $\mathcal{L}_{self}$, 
(2) stage II: fixing the weights of the VO model and fine-tuning the policy network using DD-PPO algorithm \cite{wijmans2019dd}.
Specifically, in stage II, we first load the pre-trained model provided by \cite{zhao2021surprising} for saving training time; 
then we fixate the weights of the VO and convert the output pose ($R_{t-1 \rightarrow t}$ and $T_{t-1 \rightarrow t}$) 
to the representation $P^{gps}$ needed by the policy network;
finally, we train the network using both DD-PPO \cite{wijmans2019dd} 
for updating policy network and loss $\mathcal{L}_{vc}$ for auxiliary supervision of AIM.

\textbf{Metrics.}
Three popular evaluation metrics are used: the success rate (SR), 
success weighted by path length (SPL) \cite{anderson2018evaluation} and SoftSPL \cite{datta2021integrating}.
The navigation is successful ($S=1$) when the agent takes the stop action within $0.36m$ around the target location.
SPL represents the efficiency of a navigation path 
\begin{equation}
    SPL=\frac{1}{E} \sum_{i=1}^{E} S_i \frac{l_i}{\max (l_{a_i}, l_i)},
\end{equation}
where $E$ is the total number of evaluation episodes;
$S_i \in \{0,1\}$ represents whether the agent succeeded ($S_i = 1$) in reaching the target location at the \textit{i-th} episode or not ($S_i = 0$);
$l_i$ and $l_{a_i}$ are the shortest path distances to the target location and the actual path length taken by the agent, respectively.
In addition, SoftSPL is employed to replace the binary success $S_i$ with a progress indicator ($1-\frac{d_G}{d_{init}}$) to measure how close the agent gets to
the target's global coordinate at episode termination,
where distance to goal ($d_G$) captures the geodesic distance between the agent 
and the goal upon episode termination averaged across all episodes;
$d_{init}$ is the starting geodesic distance to the goal.

\begin{table}[t]
    \centering
    \caption{Evaluation of the previous supervised methods and our model
    on the Gibson-4+ validation split. Sup. and UnSup. indicate supervised training and unsupervised training.
    SR, SPL, and SoftSPL are reported in \%.}
    \label{table:eval}
    \begin{tabular}{lcccc}
        \hline
        Method & Train & SR$\uparrow$ & SPL$\uparrow$ & SoftSPL$\uparrow$ \\
        \hline
        DeepVO \cite{wang2017deepvo}                & Sup.   & 50\scriptsize{$\pm 1$} & 39\scriptsize{$\pm 1$}  & 65\scriptsize{$\pm 0$} \\
        OA \cite{ramakrishnan2020occupancy}         & Sup.   & 38.4 & 34.3  & 40.7    \\
        SLAM-net \cite{karkus2021differentiable}    & Sup.   & 66 & 38  & -    \\
        supVO-base \cite{zhao2021surprising}        & Sup.   & 61\scriptsize{$\pm 1$} & 46\scriptsize{$\pm 1$}  & 62\scriptsize{$\pm 1$} \\
        supVO-full \cite{zhao2021surprising}        & Sup.   & 82\scriptsize{$\pm 1$} & 63\scriptsize{$\pm 1$}  & 71\scriptsize{$\pm 0$} \\
        Ours                                        & UnSup. & 71.9\scriptsize{$\pm 1.8$} & 51.2\scriptsize{$\pm 1.3$}  & 63.2\scriptsize{$\pm 0.6$} \\
        \hline
    \end{tabular}
\end{table}

\begin{table*}[t]
    \centering
    \caption{Ablation results on the Gibson-4+ validation split.
    RecInfo indicates which visual cues are used for computing re-projection loss.
    E as abbreviation for vertical edge. CP-Net and FT indicate, respectively, 
    whether the model use CP-Net with LM probability volume and fine-tune the policy network with VO.
    SR, SPL, and SoftSPL are reported in \%.}
    \label{table:ablation_eval}
    \begin{tabular}{cccccccccc}
        \hline
        Row & Visual & RecInfo & CP-Net & FT & $F_a$ & SR$\uparrow$ & SPL$\uparrow$  & SoftSPL$\uparrow$ \\
        \hline
        0   & \multicolumn{5}{c}{Ground-Truth}                  & 95.0\scriptsize{$\pm 0.7$} & 65.9\scriptsize{$\pm 0.7$}  & 65.2\scriptsize{$\pm 0.4$} \\
        \hline
        1   & RGB    & RGB     &            &            & Emb.                 & 51.0\scriptsize{$\pm 1.4$} & 36.4\scriptsize{$\pm 1.0$}  & 57.8\scriptsize{$\pm 0.7$} \\
        2   & RGBD  & RGB     &            &            & Emb.                 & 54.1\scriptsize{$\pm 2.0$} & 37.2\scriptsize{$\pm 1.4$}  & 57.5\scriptsize{$\pm 0.8$} \\
        3   & RGBD  & RGBD   &            &            & Emb.                 & 55.0\scriptsize{$\pm 1.7$} & 38.5\scriptsize{$\pm 1.0$}  & 57.7\scriptsize{$\pm 0.7$} \\
        4   & RGBD  & RGBD+E &            &            & Emb.                 & 62.1\scriptsize{$\pm 1.5$} & 42.8\scriptsize{$\pm 1.0$}  & 59.9\scriptsize{$\pm 0.7$} \\
        5   & RGBD  & RGBD+E & \checkmark &            & Emb.                 & 62.7\scriptsize{$\pm 1.6$} & 43.1\scriptsize{$\pm 1.1$}  & 59.6\scriptsize{$\pm 0.6$} \\
        6   & RGBD  & RGBD+E & \checkmark & \checkmark & Emb.                 & 65.7\scriptsize{$\pm 1.9$} & 45.3\scriptsize{$\pm 1.3$}  & 59.6\scriptsize{$\pm 0.7$} \\
        \hline
        7   & RGB    & RGB     &            & \checkmark & AIM                    & 58.5\scriptsize{$\pm 1.6$} & 41.2\scriptsize{$\pm 0.8$}  & 61.0\scriptsize{$\pm 0.8$} \\
        8   & RGBD  & RGBD   &            & \checkmark & AIM                    & 61.0\scriptsize{$\pm 1.9$} & 41.4\scriptsize{$\pm 1.3$}  & 59.1\scriptsize{$\pm 0.8$} \\
        9   & RGBD  & RGBD+E &            & \checkmark & AIM                    & 68.0\scriptsize{$\pm 2.2$} & 47.8\scriptsize{$\pm 1.7$}  & 62.2\scriptsize{$\pm 0.5$} \\
        \hline
        10  & RGBD  & RGBD+E & \checkmark & \checkmark & AIM only $\mathcal{L}_{vc}$& 67.7\scriptsize{$\pm 2.2$} & 48.1\scriptsize{$\pm 1.3$}  & 62.6\scriptsize{$\pm 0.5$} \\
        11  & RGBD  & RGBD+E & \checkmark & \checkmark & AIM w/o $\mathcal{L}_{vc}$ & 67.4\scriptsize{$\pm 2.2$} & 45.7\scriptsize{$\pm 1.8$}  & 59.6\scriptsize{$\pm 0.4$} \\
        12  & RGBD  & RGBD+E & \checkmark & \checkmark & AIM                        & 71.9\scriptsize{$\pm 1.8$} & 51.2\scriptsize{$\pm 1.3$}  & 63.2\scriptsize{$\pm 0.6$} \\
        13  & \red{RGBD+E}  & RGBD+E & \checkmark & \checkmark & AIM                & 69.2\scriptsize{$\pm 1.7$} & 48.4\scriptsize{$\pm 1.5$}  & 62.2\scriptsize{$\pm 0.4$} \\
        \hline
    \end{tabular}
\end{table*}

\begin{figure*}[t]
    \centering
    \includegraphics[width=18cm]{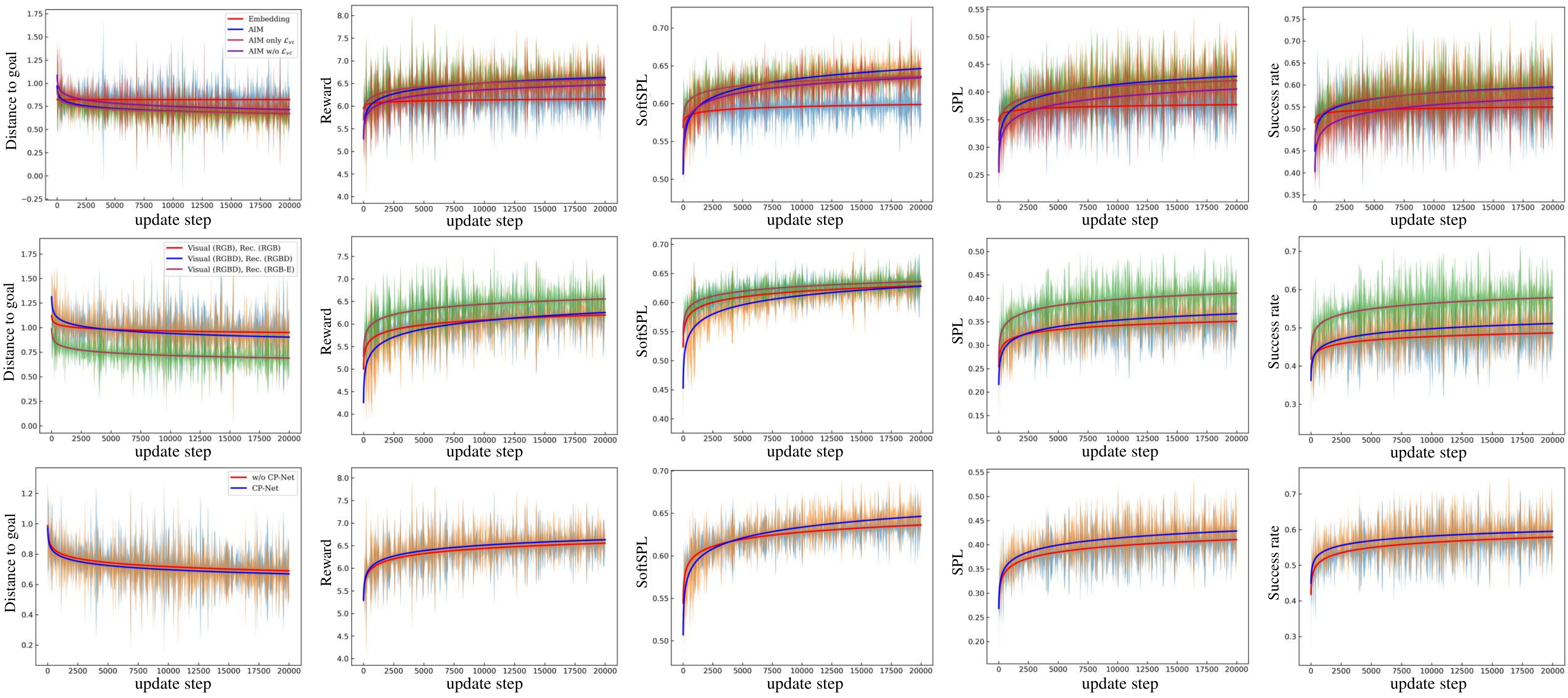}
    \caption{Plots of the distance to goal (lower is better), reward, SoftSPL, SPL and success rate (higher is better) as
    a function of the number of fine-tuning steps. 
    Top row compares the methods on how to train $F_a$, including embedding layer, AIM, AIM only $\mathcal{L}_{vc}$ and AIM w/o $\mathcal{L}_{vc}$. 
    Middle row compares the VO inputs and reconstructed error cues.
    Bottom row compares whether to use CP-Net as VO network. 
    }
    \label{fig:train_plot}
\end{figure*}

\subsection{Quantitative results}
\textbf{Comparison with state-of-the-art.} Table \ref{table:eval} shows the quantitative results of the proposed and 
compared methods on the Gibson-4+ validation set, which is a validation set of Gibson dataset according to the split used in \cite{savva2019habitat}.
Evaluation was conducted on 994 episodes from 14 validation scenes, each of which provides 71 episodes.
DeepVO \cite{wang2017deepvo} is a classical supervised RNN-based VO method.
Compared with DeepVO, our model shows an
improvement of the success rate by 43.8\% (from 50 to 71.9) and SPL by 31.3\% (from 39 to 51.2).
Occupancy Anticipation (OA) \cite{ramakrishnan2020occupancy} is a SLAM based algorithm,
with a performance of 38.4 in SR and 34.3 in SPL, which is remarkably worse than the proposed unsupervised approach.
By comparing the results of the proposed model and SLAM-net \cite{karkus2021differentiable}, 
which is a state-of-the-art SLAM based navigation method,
we can see that our algorithm has advantage for 
the success rate by 10.6\% (from 65 to 71.9) and SPL by 34.7\% (from 38 to 51.2). 
In addition, we split the supVO \cite{zhao2021surprising} into two methods for comparison, 
where supVO-base is the base model (only a network with supervised training)
and supVO-full is the full model trained with many tricks, such as action-specific design and depth top-down projection.
We can see that in terms of success rate, the proposed method outperforms the supVO-base with RGBD as inputs
by 17.9\% (from 61 to 71.9), but performs inferior to the supVO-full model.

\textbf{Ablation study.} To better understand the performance of the proposed model, 
we performed an ablation study in Table \ref{table:ablation_eval} and \ref{table:vo_errors} and Figure \ref{fig:train_plot}.
Figure \ref{fig:train_plot} plots the distance to goal, 
rewards and error metrics over time for 20000 update steps (each update step is updated after 50 time steps) 
on fine-tuning our navigation model.
Table \ref{table:ablation_eval} and \ref{table:vo_errors} show the quantitative results on navigation and VO performances.
The components to be evaluated consist of  
1) the input type (RGB or RGBD) of the VO model;
2) the visual information for computing re-projection error (RGB, RGBD or RGBD with vertical edge (RGBD+E));
3) whether or not to use the proposed CP-Net with LM probability volume;
4) approaches for encoding $F_a$, including embedding layer, AIM and training AIM with different settings;
5) whether or not to fine-tune the policy network with VO.

In this experiment, we first built a near perfect ground truth of 65.9\% SPL with 95\% success rate (Row 0 in Table \ref{table:ablation_eval}),
which was trained with perfect GPS+Compass sensors under noisy actuation.
Under the condition without fine-tuning the policy network, 
we then trained the VO models separately, and used the VO model as a drop-in replacement for ground truth
GPS without any expensive re-training.
Because such comparison is more fair, i.e., using only one variable of VO model and not involving the policy network. 
From the ablation study, we can get several conclusions as follows.

\begin{table}[t]
    \centering
    \caption{Ablation results of the proposed VO performances on the Gibson-4+ validation split.
    We report the mean absolute error (MAE) of per step for translation and rotation on polar coordinates 
    in two-dimensional space.
    }
    \label{table:vo_errors}
    \begin{tabular}{ccccc}
        \hline
        \multirow{2}{*}{Visual} & \multirow{2}{*}{RecInfo} & \multirow{2}{*}{CP-Net} & \multicolumn{2}{c}{MAE per step ($10^{-2} m$)$\downarrow$} \\
        \cmidrule(r){4-5}
         &  &  & Translation & Rotation \\
        \hline
        RGB    & RGB     &            & 6.87 \scriptsize{$\pm$13.20} & 7.59 \scriptsize{$\pm$13.64}\\
        RGBD  & RGB     &            & 5.21 \scriptsize{$\pm$9.34} & 5.83 \scriptsize{$\pm$9.93}\\
        RGBD  & RGBD   &            & 4.71 \scriptsize{$\pm$7.17} & 5.27 \scriptsize{$\pm$7.98}\\
        RGBD  & RGBD+E &            & 3.23 \scriptsize{$\pm$5.05} & 6.68 \scriptsize{$\pm$5.90}\\
        RGBD  & RGBD+E & \checkmark & 3.22 \scriptsize{$\pm$4.73} & 3.75 \scriptsize{$\pm$5.37}\\
        \red{RGBD+E}  & RGBD+E & \checkmark & 3.61 \scriptsize{$\pm$4.84} & 4.35 \scriptsize{$\pm$5.52}\\
        \hline
    \end{tabular}
\end{table}

\textit{1) Depth observation helps improve the performances of VO.}
In Table \ref{table:ablation_eval}, the RGBD sensor 
provides higher navigation success rate and SPL (Row 2, 54.1 SR, 37.2 SPL) compared with RGB-only (Row 1, 51.0 SR, 36.4 SPL).
Furthermore, with AIM (Rows 7-8), the depth observation also helps improve the performances of VO.
In Table \ref{table:vo_errors}, we can see that the VO model with depth as input 
has lower rotational and translational errors compared with RGB-only input.

\red{Note that using edge maps as additional input
leads to some degradation in the navigation performance (i.e., the last row in Table \ref{table:ablation_eval} and \ref{table:vo_errors}).
A possible reason is that the edge maps have been learned by the CNN at the early layers \cite{zeiler2014visualizing}, 
or in other words, the edge feature has been implicitly extracted by CNN. 
When we directly concatenate the edge maps as input, 
the implicit feature representation of edges may be destroyed,
which will increase the difficulty of convergence of the convolutional parameters of the early layers.}

\textit{2) Richer information for computing re-projection error is helpful.}
From Rows 2-4 in Table \ref{table:ablation_eval}, we can see that
depth sensor (Row 3), as the reconstructed visual cues, slightly improves 
the success rate by 1.7\% (from 54.1 to 55.0) and SPL by 3.5\% (from 37.2 to 38.5).
Furthermore, using texture information (vertical edge cue) for computing loss (Row 4) 
significantly improves the success rate by 12.9\% (from 55.0 to 62.1) and SPL by 11.2\% (from 38.5 to 42.8).
In the case of using AIM (Rows 8-9), the conclusion remains the same.
In addition, the direct VO performance (Table \ref{table:vo_errors}) also shows that 
training with the richer information used in this paper is more effective.

\textit{3) The proposed CP-Net with LM probability volume works better than direct pose estimation.}
Compared with Row 5 and Row 6 (SR 62.1 vs. 62.7, SPL 42.8 vs. 43.1), 
both of them have RGBD sensor as input and RGBD with edge as reconstruction target,
we can find that the VO model with CP-Net is better than the model without it (using conventional direct pose estimation \cite{godard2019digging}).
The results of Row 9 and 12 (SR 68.0 vs. 71.9, SPL 47.8 vs. 51.2) show
that the CP-Net also outperforms the direct pose estimation method when taking AIM for embedding $F_a$.

Table \ref{table:vo_errors} shows that even the MAE difference on translation prediction 
with and without CP-Net is relatively small, i.e., 3.22$\pm${\small{4.73}} vs. 3.23$\pm${\small{5.05}},  
CP-Net contributes clearly to produce lower MAE on rotation prediction with better stability (small variance) 
, i.e., 3.75$\pm${\small{5.37}} vs. 6.68$\pm${\small{5.90}}. 
Note that we only reported in Table \ref{table:vo_errors} the MAE of per step, 
the rotation and translation prediction errors accumulated over 
the entire trajectory will lead to significant differences in their final results.

In addition, the number of parameters of CP-Net (14.2 million) 
is only 9.2\% more than that of the previous method \cite{godard2019digging} (13.0 million).
Note that the proposed CP-Net is only suitable for the movements with specific actions, such as the navigation with discrete action space $A$.
Because in this case, the magnitude of the motion is limited and thus can be finely discretized. 
In more general cases (e.g., the drones with 6 degree of freedom motion), 
the cost of building LM probability volume (with more parameters and slower operation) will be large and more difficult to train.

\textit{4) Tuning policy network with VO further improves performances.}
The results in Row 6 of Table \ref{table:ablation_eval} were obtained by fixating the VO model and fine-tuning the policy network using DD-PPO algorithm \cite{wijmans2019dd}.
The whole process did not use the GPS signal. 
We can see that compared with Row 5, fine-tuning the policy network further improves
the success rate by 4.8\% (from 62.7 to 65.7) and SPL by 5.1\% (from 43.1 to 45.3).

\textit{5) AIM significantly improves the performances compared with action embedding layer.}
Compared with Row 1 and Row 7 in Table \ref{table:ablation_eval}(14.7\% improvement in SR), Row 3 and Row 8 (10.9\% improvement in SR), 
Row 4 and Row 9 (9.5\% improvement in SR), Row 6 and Row 12 (9.4\% improvement in SR),
we can clearly see that using AIM to encode $F_a$ significantly improves the navigation performances.
In addition, it is clear that the lower the navigation capability (the less accurate the VO) is, 
the larger performance boost AIM can give.

\textit{6) The visual correction loss $\mathcal{L}_{vc}$ is useful for training AIM.}
The results of Row 10 and Row 11 in Table \ref{table:ablation_eval} show incomplete training of AIM.
We can find that without $\mathcal{L}_{vc}$ (only receives the policy gradient),
the navigation performances are improved (Row 6 vs. Row 11), but far inferior to that of the AIM with $\mathcal{L}_{vc}$ (Row 12).
Interestingly, detaching the AIM and training it only with $\mathcal{L}_{vc}$ (Row 10) can still improve the performances compared with embedding layer (Row 6).
This finding may be a side-effect of the fact that encoding position and head direction with neural representation 
helps improve the performances of navigation.

\subsection{Performances on the noise condition}

\begin{figure}[t]
    \centering
    \includegraphics[width=8cm]{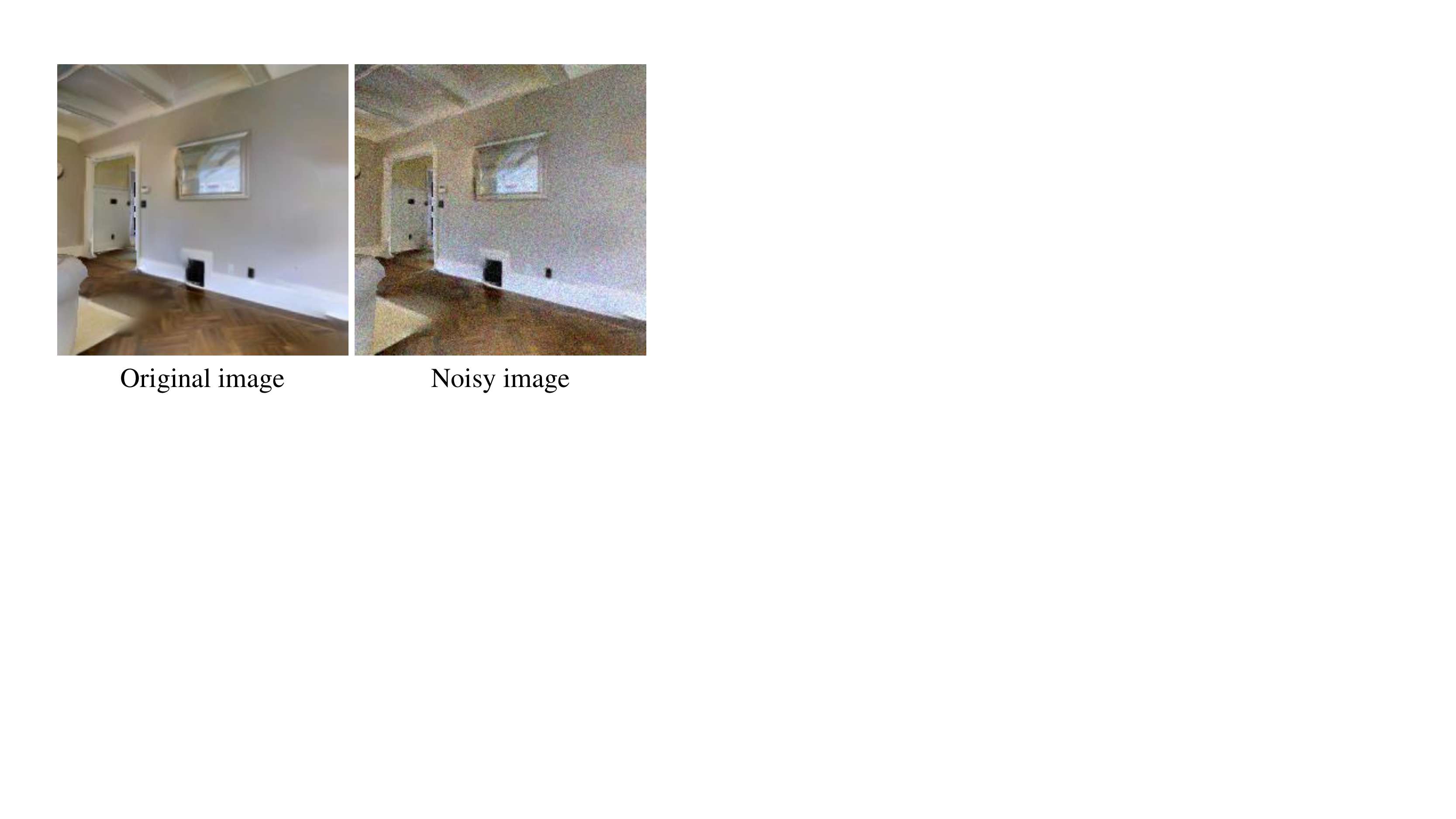}
    \caption{
        Visualization of a scene in Gibson-4+ dataset.
        The images from left to right are the original and the one added with Gaussian noise ($\sigma=0.05$), respectively.
    }
    
    \label{fig:vis_noise}
\end{figure}

\begin{table}[t]
    \centering
    \caption{The effect of noise on the proposed model. SR, SPL, and SoftSPL are reported in \%.
    \red{Symbols DN, RGBN and DM indicate depth noise, RGB noise (indicated by the variance $\sigma$) and denoising method, respectively.
    "Gaussian" and "Bilateral" denote respectively the Gaussian and bilateral filtering algorithms.}}
    
    \label{table:noise_eval}
    \begin{tabular}{cccccc}
        \hline
        DN & RGBN & DM & SR$\uparrow$ & SPL$\uparrow$  & SoftSPL$\uparrow$ \\
        \hline
                   &      &           & 71.9\scriptsize{$\pm 1.8$} & 51.2\scriptsize{$\pm 1.3$} & 63.2\scriptsize{$\pm 0.6$} \\
                   & \red{0.01} &           & 68.4\scriptsize{$\pm 1.8$} & 48.4\scriptsize{$\pm 1.4$} & 61.8\scriptsize{$\pm 0.3$} \\
                   & \red{0.03} &           & 61.8\scriptsize{$\pm 2.0$} & 44.3\scriptsize{$\pm 1.5$} & 61.1\scriptsize{$\pm 0.5$} \\
                   & 0.05 &           & 41.7\scriptsize{$\pm 2.1$} & 29.5\scriptsize{$\pm 1.7$} & 57.4\scriptsize{$\pm 0.3$} \\
                   & 0.05 & \red{Gaussian}  & 67.6\scriptsize{$\pm 1.7$} & 47.2\scriptsize{$\pm 1.4$} & 60.9\scriptsize{$\pm 0.7$} \\
                   & 0.05 & \red{Bilateral} & 66.8\scriptsize{$\pm 1.7$} & 47.1\scriptsize{$\pm 1.3$} & 61.7\scriptsize{$\pm 0.7$} \\
        \checkmark &      &           & 59.1\scriptsize{$\pm 3.0$} & 40.8\scriptsize{$\pm 1.9$} & 59.5\scriptsize{$\pm 0.6$} \\
        \checkmark & 0.05 &           & 40.8\scriptsize{$\pm 1.9$} & 29.6\scriptsize{$\pm 1.2$} & 59.1\scriptsize{$\pm 0.6$} \\
        
        \hline
    \end{tabular}
\end{table}

In real environments, sensors, especially depth sensors, may be affected by noise, which may degrade the navigation performance.
We report the performance of the proposed method on the noisy environments in Table \ref{table:noise_eval}. 
The RGB image is added with Gaussian noise with a variance of \red{$\sigma=0.01, 0.03, 0.05$ }
and the depth noise is added using redwood noise model, like \cite{zhao2021surprising}.
As shown in Figure \ref{fig:vis_noise}, 
the quality of the original images in the simulated dataset Gibson-4+ used in this paper 
is not quite high due to the existence of detail blurring, 
and after being added with Gaussian noise with $\sigma=0.05$, the image is significantly degraded. 
This noise intensity ($\sigma=0.05$) added in this experiment is relatively large for practical scenes \cite{abdelhamed2018high}.

It is clear that the reprojection error that depends on the consistency of re-projection with the vision is sensitive to noise.
This performance is expected because in the framework of unsupervised learning, 
the reprojection error is correctly predicated on static scenes.
Thus, the loss is difficult to converge when the input noise is severe.
In Table \ref{table:noise_eval}, we can find \red{several} phenomena from the results of two noisy conditions, i.e., 
redwood noise model for depth and Gaussian noise for RGB image.
\red{
Firstly, depth noise has less effect than RGB noise with $\sigma=0.05$ on navigational metrics.
In Table \ref{table:noise_eval}, the success rate of navigation performance achieves 59.1\% when introducing only the depth noise,
while the performance is 41.7\% when containing the RGB noise with $\sigma=0.05$.
Secondly, higher intensity of the RGB noise lead to greater effect on the results.
For example, when the noise variance $\sigma$ increases from 0 to 0.01, 
the degradation of success rate is 5.1\% (reduced from 71.9\% to 68.4\%);
when $\sigma$ increases from 0.01 to 0.03, the degradation of success rate is 10.7\% (from 68.4\% to 61.8\%)
and when $\sigma$ increases from 0.03 to 0.05, the degradation of success rate is 48.2\% (from 61.8\% to 41.7\%).
Finally, simple denoising algorithms, such as Gaussian and bilateral filtering, can effectively improve the navigation performance.
In Table \ref{table:noise_eval}, 
we can see that these two filters can greatly reduce the effects of noise.
}


\begin{figure}[t]
    \centering
    \includegraphics[width=8cm]{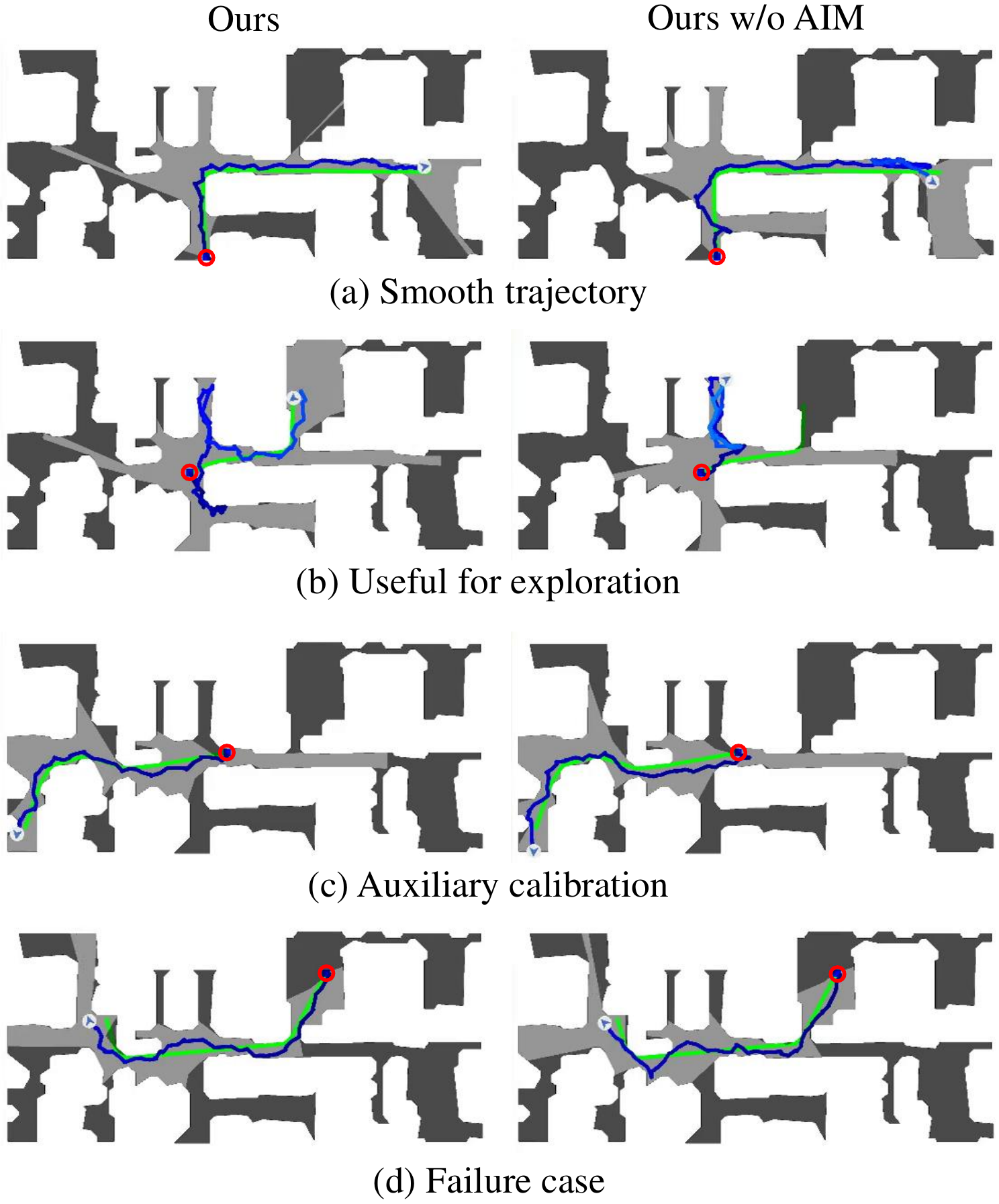}
    \caption{Visualization of four PointGoal navigation episodes in the \textit{Cantwell} scene.
    Red circle, green line and blue line indicate starting point, ground truth trajectory and predicted action trajectory, respectively.
    The position and direction of blue arrow indicate the position and orientation of the robot in its final time step.}
    \label{fig:vis_example}
\end{figure}

\begin{figure*}[t]
    \centering
    \includegraphics[width=18cm]{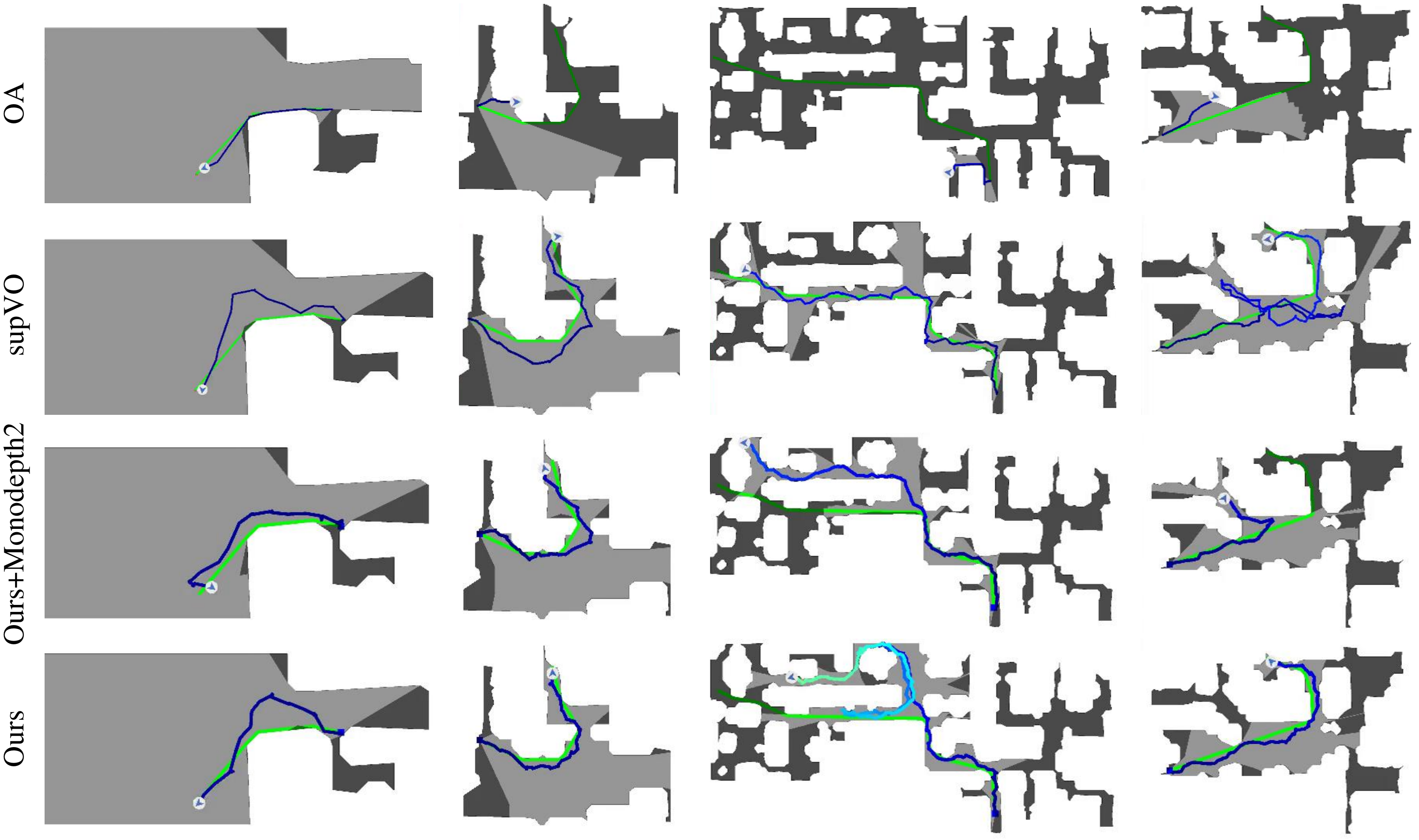}
    \caption{
    Visualization of the results by the proposed and other navigation methods.
    The rows from the top to bottom are Occupancy Anticipation (OA) \cite{ramakrishnan2020occupancy}, 
    supVO \cite{zhao2021surprising}, 
    the approach replacing the VO module of proposed navigation model with Monodepth2 \cite{Godard-iccv19},
    and the proposed model, respectively.
    Green line and blue line indicate ground truth trajectory and predicted action trajectory, respectively.
    The blue line fading means the number of agent's actions is greater than 200.
}
    \label{fig:vis_example_sota}
\end{figure*}

\subsection{Qualitative results}
We show four typical examples of episodes for PointGoal navigation in Figure \ref{fig:vis_example}.
The first three rows show success cases compared with the method without AIM (Figure \ref{fig:vis_example}(a-c)) .
From the navigation results for each episode, 
we can find several typical advantages of AIM, 
including the ability to better smooth the trajectory of the agent (Figure \ref{fig:vis_example}(a)), 
to better facilitate exploration (Figure \ref{fig:vis_example}(b)), 
and to help the intelligent agent calibrate errors in the visual signal (Figure \ref{fig:vis_example}(c)).
We also show a failure case in Figure \ref{fig:vis_example}(d).
This is a typical failure case that the agent incorrectly estimates the endpoint due to the accumulated large error by VO
even our model with AIM can find a trajectory that is relatively closer to the ground truth.

Figure \ref{fig:vis_example_sota} shows the navigation results of different algorithms.
As a SLAM based algorithm, OA \cite{ramakrishnan2020occupancy} uses a planning method to drive the agent to the target location.
However, the method performs poorly on global optimization 
but shows good performance when there are few obstacles between the initial location and the target (the first column).
SupVO \cite{zhao2021surprising} is a reinforcement learning based approach and uses 
supervised learning to train a precise VO model. 
Compared with our model, supVO performs well in long-distance navigation (the third column).
The proposed approach tends to go to a wrong endpoint in long-distance navigation due to the VO accuracy.
We also compared with recent VO method Monodepth2 \cite{Godard-iccv19} in navigation task.
As can be seen, the method Ours+Monodepth2 which uses Monodepth2 as the VO module 
generally performs worse than the proposed model.

\section{Conclusions}
This paper demonstrates that it is feasible to train a policy network without using GPS signals in indoor environment.
The main solution is to use VO and action integration.
We use an unsupervised VO algorithm and futher improve the performance by 
1) better reconstructing the target image with richer visual cues, e.g., depth and edge,
and 2) proposing a CP-Net with LM probability volume using pose discretization manner.
In addition, we propose the AIM to predict action-only odometry using neural representations.
The results show that the proposed method achieves satisfactory results and 
outperforms the state-of-the-art partially supervised learning algorithms on the popular Gibson dataset.
Furthermore, the AIM is demonstrated 
to help improve the navigation ability of the intelligent agent under inaccurate positional estimation.

The limitation of the proposed VO algorithm is that only static scene is considered and not robust enough for RGB noise due to reprojection error.
In addition, our algorithm was tested only on the simulated 3D indoor environments,
which may not verify the domain adaptation capability, especially on the real world environments. 
Therefore, our future research will focus on testing and refining the algorithm on a wider range of data and realistic environments.

\bibliographystyle{IEEEtran}


\begin{thebibliography}{10}
\providecommand{\url}[1]{#1}
\csname url@samestyle\endcsname
\providecommand{\newblock}{\relax}
\providecommand{\bibinfo}[2]{#2}
\providecommand{\BIBentrySTDinterwordspacing}{\spaceskip=0pt\relax}
\providecommand{\BIBentryALTinterwordstretchfactor}{4}
\providecommand{\BIBentryALTinterwordspacing}{\spaceskip=\fontdimen2\font plus
\BIBentryALTinterwordstretchfactor\fontdimen3\font minus
  \fontdimen4\font\relax}
\providecommand{\BIBforeignlanguage}[2]{{%
\expandafter\ifx\csname l@#1\endcsname\relax
\typeout{** WARNING: IEEEtran.bst: No hyphenation pattern has been}%
\typeout{** loaded for the language `#1'. Using the pattern for}%
\typeout{** the default language instead.}%
\else
\language=\csname l@#1\endcsname
\fi
#2}}
\providecommand{\BIBdecl}{\relax}
\BIBdecl

\bibitem{wijmans2019dd}
E.~Wijmans, A.~Kadian, A.~Morcos, S.~Lee, I.~Essa, D.~Parikh, M.~Savva, and
  D.~Batra, ``Dd-ppo: Learning near-perfect pointgoal navigators from 2.5
  billion frames,'' in \emph{International Conference on Learning
  Representations}, 2019.

\bibitem{zhao2021surprising}
X.~Zhao, H.~Agrawal, D.~Batra, and A.~G. Schwing, ``The surprising
  effectiveness of visual odometry techniques for embodied pointgoal
  navigation,'' in \emph{Proceedings of the IEEE/CVF International Conference
  on Computer Vision}, 2021, pp. 16\,127--16\,136.

\bibitem{karkus2021differentiable}
P.~Karkus, S.~Cai, and D.~Hsu, ``Differentiable slam-net: Learning particle
  slam for visual navigation,'' in \emph{Proceedings of the IEEE/CVF Conference
  on Computer Vision and Pattern Recognition}, 2021, pp. 2815--2825.

\bibitem{mandal2018animals}
S.~Mandal, ``How do animals find their way back home? a brief overview of
  homing behavior with special reference to social hymenoptera,''
  \emph{Insectes sociaux}, vol.~65, no.~4, pp. 521--536, 2018.

\bibitem{o1978hippocampal}
J.~O'Keefe and D.~H. Conway, ``Hippocampal place units in the freely moving
  rat: why they fire where they fire,'' \emph{Experimental brain research},
  vol.~31, no.~4, pp. 573--590, 1978.

\bibitem{taube1990head}
J.~S. Taube, R.~U. Muller, and J.~B. Ranck, ``Head-direction cells recorded
  from the postsubiculum in freely moving rats. i. description and quantitative
  analysis,'' \emph{Journal of Neuroscience}, vol.~10, no.~2, pp. 420--435,
  1990.

\bibitem{hafting2005microstructure}
T.~Hafting, M.~Fyhn, S.~Molden, M.-B. Moser, and E.~I. Moser, ``Microstructure
  of a spatial map in the entorhinal cortex,'' \emph{Nature}, vol. 436, no.
  7052, pp. 801--806, 2005.

\bibitem{etienne1996path}
A.~S. Etienne, R.~Maurer, and V.~S{\'e}guinot, ``Path integration in mammals
  and its interaction with visual landmarks.'' \emph{The Journal of
  experimental biology}, vol. 199, no.~1, pp. 201--209, 1996.

\bibitem{zhou2017unsupervised}
T.~Zhou, M.~Brown, N.~Snavely, and D.~G. Lowe, ``Unsupervised learning of depth
  and ego-motion from video,'' in \emph{Proceedings of the IEEE conference on
  computer vision and pattern recognition}, 2017, pp. 1851--1858.

\bibitem{wang20223d}
G.~Wang, J.~Zhong, S.~Zhao, W.~Wu, Z.~Liu, and H.~Wang, ``3d hierarchical
  refinement and augmentation for unsupervised learning of depth and pose from
  monocular video,'' \emph{IEEE Transactions on Circuits and Systems for Video
  Technology}, 2022.

\bibitem{hochreiter1997long}
S.~Hochreiter and J.~Schmidhuber, ``Long short-term memory,'' \emph{Neural
  computation}, vol.~9, no.~8, pp. 1735--1780, 1997.

\bibitem{banino2018}
A.~Banino, C.~Barry, B.~Uria, C.~Blundell, T.~Lillicrap, P.~Mirowski,
  A.~Pritzel, M.~J. Chadwick, T.~Degris, J.~Modayil \emph{et~al.},
  ``Vector-based navigation using grid-like representations in artificial
  agents,'' \emph{Nature}, vol. 557, no. 7705, pp. 429--433, 2018.

\bibitem{moser2014grid}
E.~I. Moser, Y.~Roudi, M.~P. Witter, C.~Kentros, T.~Bonhoeffer, and M.-B.
  Moser, ``Grid cells and cortical representation,'' \emph{Nature Reviews
  Neuroscience}, vol.~15, no.~7, pp. 466--481, 2014.

\bibitem{savva2019habitat}
M.~Savva, A.~Kadian, O.~Maksymets, Y.~Zhao, E.~Wijmans, B.~Jain, J.~Straub,
  J.~Liu, V.~Koltun, J.~Malik \emph{et~al.}, ``Habitat: A platform for embodied
  ai research,'' in \emph{Proceedings of the IEEE/CVF International Conference
  on Computer Vision}, 2019, pp. 9339--9347.

\bibitem{xia2018gibson}
F.~Xia, A.~R. Zamir, Z.~He, A.~Sax, J.~Malik, and S.~Savarese, ``Gibson env:
  Real-world perception for embodied agents,'' in \emph{Proceedings of the IEEE
  conference on computer vision and pattern recognition}, 2018, pp. 9068--9079.

\bibitem{he2016deep}
K.~He, X.~Zhang, S.~Ren, and J.~Sun, ``Deep residual learning for image
  recognition,'' in \emph{Proceedings of the IEEE conference on computer vision
  and pattern recognition}, 2016, pp. 770--778.

\bibitem{thrun1998learning}
S.~Thrun, ``Learning metric-topological maps for indoor mobile robot
  navigation,'' \emph{Artificial Intelligence}, vol.~99, no.~1, pp. 21--71,
  1998.

\bibitem{ding2022monocular}
J.~Ding, L.~Gao, W.~Liu, H.~Piao, J.~Pan, Z.~Du, X.~Yang, and B.~Yin,
  ``Monocular camera-based complex obstacle avoidance via efficient deep
  reinforcement learning,'' \emph{IEEE Transactions on Circuits and Systems for
  Video Technology}, 2022.

\bibitem{chaplot2019learning}
D.~S. Chaplot, D.~Gandhi, S.~Gupta, A.~Gupta, and R.~Salakhutdinov, ``Learning
  to explore using active neural slam,'' in \emph{International Conference on
  Learning Representations}, 2019.

\bibitem{kwon2021visual}
O.~Kwon, N.~Kim, Y.~Choi, H.~Yoo, J.~Park, and S.~Oh, ``Visual graph memory
  with unsupervised representation for visual navigation,'' in
  \emph{Proceedings of the IEEE/CVF International Conference on Computer
  Vision}, 2021, pp. 15\,890--15\,899.

\bibitem{du2020vtnet}
H.~Du, X.~Yu, and L.~Zheng, ``Vtnet: Visual transformer network for object goal
  navigation,'' in \emph{International Conference on Learning Representations},
  2020.

\bibitem{zhang2021hierarchical}
S.~Zhang, X.~Song, Y.~Bai, W.~Li, Y.~Chu, and S.~Jiang, ``Hierarchical
  object-to-zone graph for object navigation,'' in \emph{Proceedings of the
  IEEE/CVF International Conference on Computer Vision}, 2021, pp.
  15\,130--15\,140.

\bibitem{zhang2020language}
W.~Zhang, C.~Ma, Q.~Wu, and X.~Yang, ``Language-guided navigation via
  cross-modal grounding and alternate adversarial learning,'' \emph{IEEE
  Transactions on Circuits and Systems for Video Technology}, vol.~31, no.~9,
  pp. 3469--3481, 2020.

\bibitem{gupta2017cognitive}
S.~Gupta, J.~Davidson, S.~Levine, R.~Sukthankar, and J.~Malik, ``Cognitive
  mapping and planning for visual navigation,'' in \emph{Proceedings of the
  IEEE Conference on Computer Vision and Pattern Recognition}, 2017, pp.
  2616--2625.

\bibitem{mayo2021visual}
B.~Mayo, T.~Hazan, and A.~Tal, ``Visual navigation with spatial attention,'' in
  \emph{Proceedings of the IEEE/CVF Conference on Computer Vision and Pattern
  Recognition}, 2021, pp. 16\,898--16\,907.

\bibitem{ye2021auxiliary}
J.~Ye, D.~Batra, A.~Das, and E.~Wijmans, ``Auxiliary tasks and exploration
  enable objectgoal navigation,'' in \emph{Proceedings of the IEEE/CVF
  International Conference on Computer Vision}, 2021, pp. 16\,117--16\,126.

\bibitem{maksymets2021thda}
O.~Maksymets, V.~Cartillier, A.~Gokaslan, E.~Wijmans, W.~Galuba, S.~Lee, and
  D.~Batra, ``Thda: Treasure hunt data augmentation for semantic navigation,''
  in \emph{Proceedings of the IEEE/CVF International Conference on Computer
  Vision}, 2021, pp. 15\,374--15\,383.

\bibitem{deng2020evolving}
Z.~Deng, K.~Narasimhan, and O.~Russakovsky, ``Evolving graphical planner:
  Contextual global planning for vision-and-language navigation,''
  \emph{Advances in Neural Information Processing Systems}, vol.~33, pp.
  20\,660--20\,672, 2020.

\bibitem{chen2021topological}
K.~Chen, J.~K. Chen, J.~Chuang, M.~V{\'a}zquez, and S.~Savarese, ``Topological
  planning with transformers for vision-and-language navigation,'' in
  \emph{Proceedings of the IEEE/CVF Conference on Computer Vision and Pattern
  Recognition}, 2021, pp. 11\,276--11\,286.

\bibitem{georgakis2022cross}
G.~Georgakis, K.~Schmeckpeper, K.~Wanchoo, S.~Dan, E.~Miltsakaki, D.~Roth, and
  K.~Daniilidis, ``Cross-modal map learning for vision and language
  navigation,'' in \emph{Proceedings of the IEEE/CVF Conference on Computer
  Vision and Pattern Recognition}, 2022, pp. 15\,460--15\,470.

\bibitem{Engel-pami18}
J.~{Engel}, V.~{Koltun}, and D.~{Cremers}, ``Direct sparse odometry,''
  \emph{{IEEE} Trans. Pattern Anal. Mach. Intell.}, vol.~40, no.~3, pp.
  611--625, 2018.

\bibitem{Mur-tro17}
R.~Mur-Artal and J.~D. Tard\'os, ``{ORB-SLAM2}: an open-source {SLAM} system
  for monocular, stereo and {RGB-D} cameras,'' \emph{{IEEE} Trans. Robot.},
  vol.~33, no.~5, pp. 1255--1262, 2017.

\bibitem{Godard-iccv19}
C.~Godard, O.~{Mac Aodha}, M.~Firman, and G.~J. Brostow, ``Digging into
  self-supervised monocular depth prediction,'' in \emph{ICCV}, 2019.

\bibitem{Wang-icra17}
S.~Wang, R.~Clark, H.~Wen, and N.~Trigoni, ``Deepvo: Towards end-to-end visual
  odometry with deep recurrent convolutional neural networks,'' in \emph{ICRA},
  2017.

\bibitem{Xue-cvpr19}
F.~Xue, X.~Wang, S.~Li, Q.~Wang, J.~Wang, and H.~Zha, ``Beyond tracking:
  Selecting memory and refining poses for deep visual odometry,'' in
  \emph{CVPR}, 2019.

\bibitem{Tang-iclr19}
C.~Tang and P.~Tan, ``Ba-net: Dense bundle adjustment network,'' in
  \emph{ICLR}, 2019.

\bibitem{Zhou-cvpr17}
T.~Zhou, M.~Brown, N.~Snavely, and D.~G. Lowe, ``Unsupervised learning of depth
  and ego-motion from video,'' in \emph{CVPR}, 2017.

\bibitem{Zou-eccv20}
Y.~Zou, P.~Ji, Q.-H. Tran, J.-B. Huang, and M.~Chandraker, ``Learning monocular
  visual odometry via self-supervised long-term modeling,'' in \emph{ECCV},
  2020.

\bibitem{Zhan-icra19}
H.~Zhan, C.~S. Weerasekera, J.~Bian, and I.~Reid, ``Visual odometry revisited:
  What should be learnt?'' in \emph{ICRA}, 2019.

\bibitem{cao2023learning}
Y.-J. Cao, X.-S. Zhang, F.-Y. Luo, P.~Peng, C.~Lin, K.-F. Yang, and Y.-J. Li,
  ``Learning generalized visual odometry using position-aware optical flow and
  geometric bundle adjustment,'' \emph{Pattern Recognition}, vol. 136, p.
  109262, 2023.

\bibitem{9386100}
S.~Chen, Z.~Pu, X.~Fan, and B.~Zou, ``Fixing defect of photometric loss for
  self-supervised monocular depth estimation,'' \emph{IEEE Transactions on
  Circuits and Systems for Video Technology}, vol.~32, no.~3, pp. 1328--1338,
  2022.

\bibitem{wei2021iterative}
Y.~Wei, H.~Guo, J.~Lu, and J.~Zhou, ``Iterative feature matching for
  self-supervised indoor depth estimation,'' \emph{IEEE Transactions on
  Circuits and Systems for Video Technology}, 2021.

\bibitem{tian2021depth}
F.~Tian, Y.~Gao, Z.~Fang, Y.~Fang, J.~Gu, H.~Fujita, and J.-N. Hwang, ``Depth
  estimation using a self-supervised network based on cross-layer feature
  fusion and the quadtree constraint,'' \emph{IEEE transactions on circuits and
  systems for video technology}, 2021.

\bibitem{song2021monocular}
M.~Song, S.~Lim, and W.~Kim, ``Monocular depth estimation using laplacian
  pyramid-based depth residuals,'' \emph{IEEE transactions on circuits and
  systems for video technology}, vol.~31, no.~11, pp. 4381--4393, 2021.

\bibitem{ioffe2015batch}
S.~Ioffe and C.~Szegedy, ``Batch normalization: Accelerating deep network
  training by reducing internal covariate shift,'' in \emph{International
  conference on machine learning}.\hskip 1em plus 0.5em minus 0.4em\relax PMLR,
  2015, pp. 448--456.

\bibitem{wu2018group}
Y.~Wu and K.~He, ``Group normalization,'' in \emph{Proceedings of the European
  conference on computer vision (ECCV)}, 2018, pp. 3--19.

\bibitem{godard2019digging}
C.~Godard, O.~Mac~Aodha, M.~Firman, and G.~J. Brostow, ``Digging into
  self-supervised monocular depth estimation,'' in \emph{Proceedings of the
  IEEE/CVF International Conference on Computer Vision}, 2019, pp. 3828--3838.

\bibitem{Zhou-tip04}
{Zhou Wang}, A.~C. {Bovik}, H.~R. {Sheikh}, and E.~P. {Simoncelli}, ``Image
  quality assessment: from error visibility to structural similarity,''
  \emph{{IEEE} Trans. Image Process.}, vol.~13, no.~4, pp. 600--612, 2004.

\bibitem{hubel1962receptive}
D.~H. Hubel and T.~N. Wiesel, ``Receptive fields, binocular interaction and
  functional architecture in the cat's visual cortex,'' \emph{The Journal of
  physiology}, vol. 160, no.~1, p. 106, 1962.

\bibitem{zeng2011center}
C.~Zeng, Y.~Li, and C.~Li, ``Center--surround interaction with adaptive
  inhibition: A computational model for contour detection,'' \emph{NeuroImage},
  vol.~55, no.~1, pp. 49--66, 2011.

\bibitem{cao2019application}
Y.-J. Cao, C.~Lin, Y.-J. Pan, and H.-J. Zhao, ``Application of the
  center--surround mechanism to contour detection,'' \emph{Multimedia Tools and
  Applications}, vol.~78, no.~17, pp. 25\,121--25\,141, 2019.

\bibitem{chen2018progressively}
H.~Chen and Y.~Li, ``Progressively complementarity-aware fusion network for
  rgb-d salient object detection,'' in \emph{CVPR}, 2018, pp. 3051--3060.

\bibitem{caolearning}
Y.-J. Cao, C.~Lin, and Y.-J. Li, ``Learning crisp boundaries using deep
  refinement network and adaptive weighting loss,'' \emph{IEEE Transactions on
  Multimedia}, vol.~23, pp. 761--771, 2021.

\bibitem{taube1998}
J.~S. Taube, ``Head direction cells and the neurophysiological basis for a
  sense of direction,'' \emph{Progress in neurobiology}, vol.~55, no.~3, pp.
  225--256, 1998.

\bibitem{moser2008}
E.~I. Moser, E.~Kropff, and M.-B. Moser, ``Place cells, grid cells, and the
  brain's spatial representation system,'' \emph{Annu. Rev. Neurosci.},
  vol.~31, pp. 69--89, 2008.

\bibitem{murali2019pyrobot}
A.~Murali, T.~Chen, K.~V. Alwala, D.~Gandhi, L.~Pinto, S.~Gupta, and A.~Gupta,
  ``Pyrobot: An open-source robotics framework for research and benchmarking,''
  \emph{arXiv preprint arXiv:1906.08236}, 2019.

\bibitem{paszke2017automatic}
A.~Paszke, S.~Gross, S.~Chintala, G.~Chanan, E.~Yang, Z.~DeVito, Z.~Lin,
  A.~Desmaison, L.~Antiga, and A.~Lerer, ``Automatic differentiation in
  pytorch,'' 2017.

\bibitem{kingma2014adam}
D.~P. Kingma and J.~Ba, ``Adam: A method for stochastic optimization,''
  \emph{arXiv preprint arXiv:1412.6980}, 2014.

\bibitem{anderson2018evaluation}
P.~Anderson, A.~Chang, D.~S. Chaplot, A.~Dosovitskiy, S.~Gupta, V.~Koltun,
  J.~Kosecka, J.~Malik, R.~Mottaghi, M.~Savva \emph{et~al.}, ``On evaluation of
  embodied navigation agents,'' \emph{arXiv preprint arXiv:1807.06757}, 2018.

\bibitem{datta2021integrating}
S.~Datta, O.~Maksymets, J.~Hoffman, S.~Lee, D.~Batra, and D.~Parikh,
  ``Integrating egocentric localization for more realistic point-goal
  navigation agents,'' in \emph{Conference on Robot Learning}.\hskip 1em plus
  0.5em minus 0.4em\relax PMLR, 2021, pp. 313--328.

\bibitem{wang2017deepvo}
S.~Wang, R.~Clark, H.~Wen, and N.~Trigoni, ``Deepvo: Towards end-to-end visual
  odometry with deep recurrent convolutional neural networks,'' in \emph{2017
  IEEE international conference on robotics and automation (ICRA)}.\hskip 1em
  plus 0.5em minus 0.4em\relax IEEE, 2017, pp. 2043--2050.

\bibitem{ramakrishnan2020occupancy}
S.~K. Ramakrishnan, Z.~Al-Halah, and K.~Grauman, ``Occupancy anticipation for
  efficient exploration and navigation,'' in \emph{ECCV}.\hskip 1em plus 0.5em
  minus 0.4em\relax Springer, 2020, pp. 400--418.

\bibitem{zeiler2014visualizing}
M.~D. Zeiler and R.~Fergus, ``Visualizing and understanding convolutional
  networks,'' in \emph{ECCV}.\hskip 1em plus 0.5em minus 0.4em\relax Springer,
  2014, pp. 818--833.

\bibitem{abdelhamed2018high}
A.~Abdelhamed, S.~Lin, and M.~S. Brown, ``A high-quality denoising dataset for
  smartphone cameras,'' in \emph{CVPR}, 2018, pp. 1692--1700.

\end{thebibliography}

\begin{IEEEbiography}[{\includegraphics[width=1in,height=1.25in,clip,keepaspectratio]{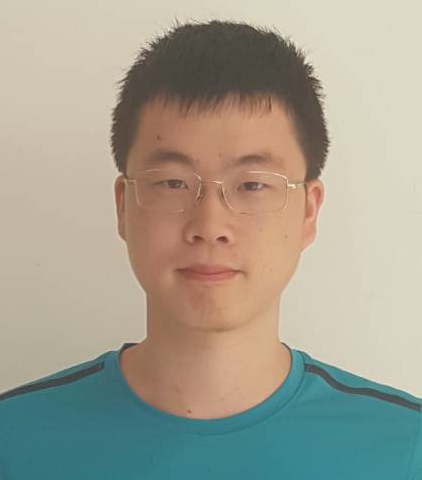}}]
{Yijun Cao} received the M.S. degree from the
College of Electric and Information Engineering,
Guangxi University of Science and Technology. Currently, he is working toward the Ph.D. degree with
the College of Electric and Information Engineering,
University of Electronic Science and Technology of
China (UESTC). His area of research are visual SLAM and navigation.
\end{IEEEbiography}
\vspace{11pt}

\begin{IEEEbiography}[{\includegraphics[width=1in,height=1.25in,clip,keepaspectratio]{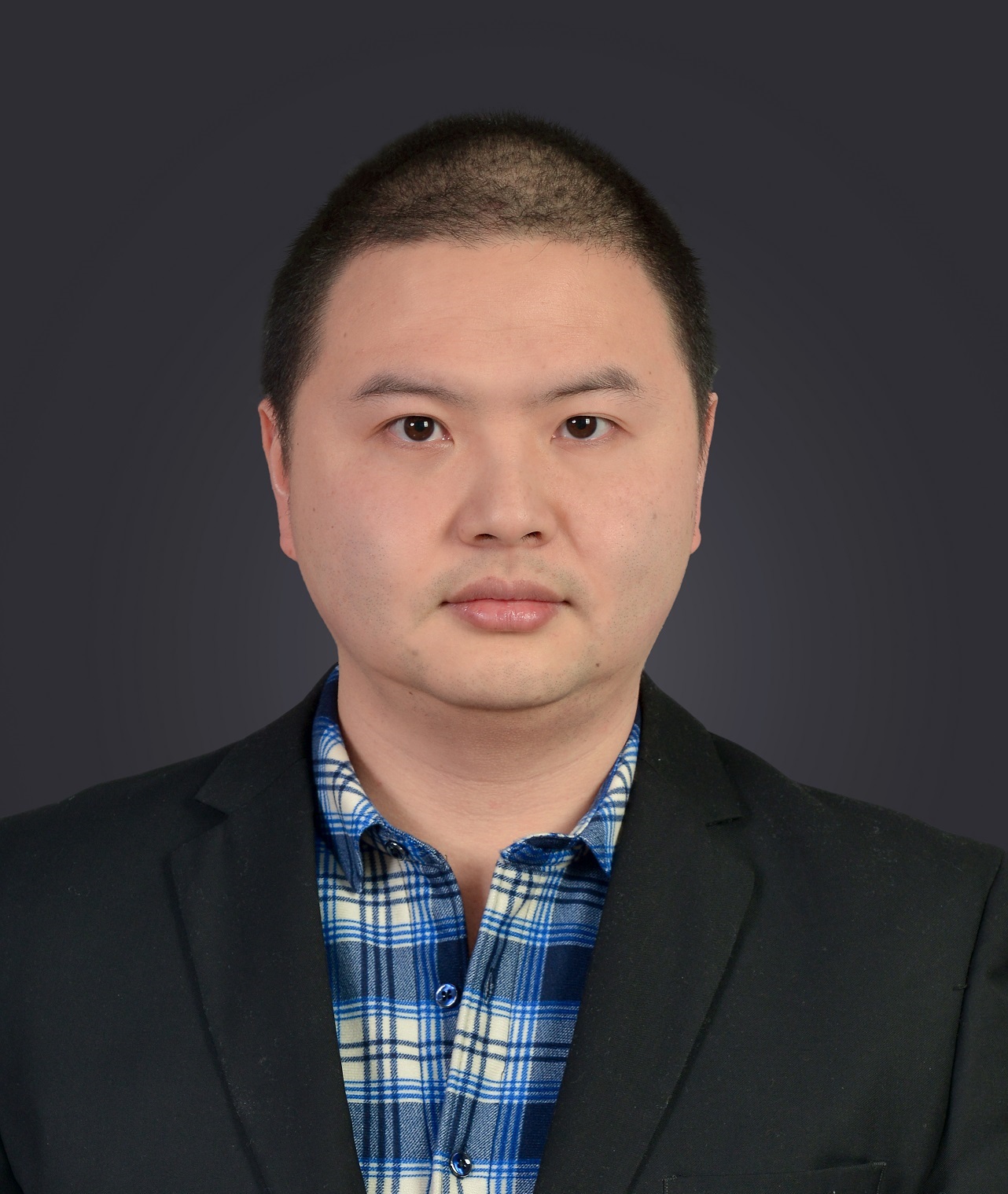}}]
{Xian-Shi Zhang} received the Ph.D. degree in biomedical engineering from the University of Electronic Science and Technology of China (UESTC), 
Chengdu, China, in 2017. He is currently an Assistant Research Professor with the MOE Key Laboratory for Neuroinformation, 
School of Life Science and Technology, UESTC. 
His research interests include visual mechanism modeling and bio-inspired computer vision.
\end{IEEEbiography}
\vspace{11pt}

\begin{IEEEbiography}[{\includegraphics[width=1in,height=1.25in,clip,keepaspectratio]{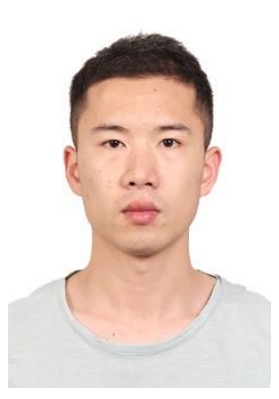}}]
{Fuya Luo} received the B.S. degree in biomedical engineering from University of Electronic Science and Technology of China (UESTC), in 2015. He is now
pursuing his Ph.D. degree in UESTC. His research interests include scene understanding, brain-inspired computer vision, weakly supervised learning, and
image-to-image translation.
\end{IEEEbiography}
\vspace{11pt}

\begin{IEEEbiography}[{\includegraphics[width=1in,height=1.25in,clip,keepaspectratio]{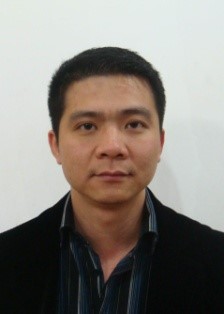}}]
{Chuan Lin} Chuan Lin received the Ph.D. degree from the Nanjing University of Aeronautics and Astronautics,
China, in 2019. He is a Professor with the College of Electric and Information Engineering, 
Guangxi University of Science and Technology, China. His areas
of research are bionic intelligent computing, computer vision, and pattern recognition.
\end{IEEEbiography}
\vspace{11pt}

\begin{IEEEbiography}[{\includegraphics[width=1in,height=1.25in,clip,keepaspectratio]{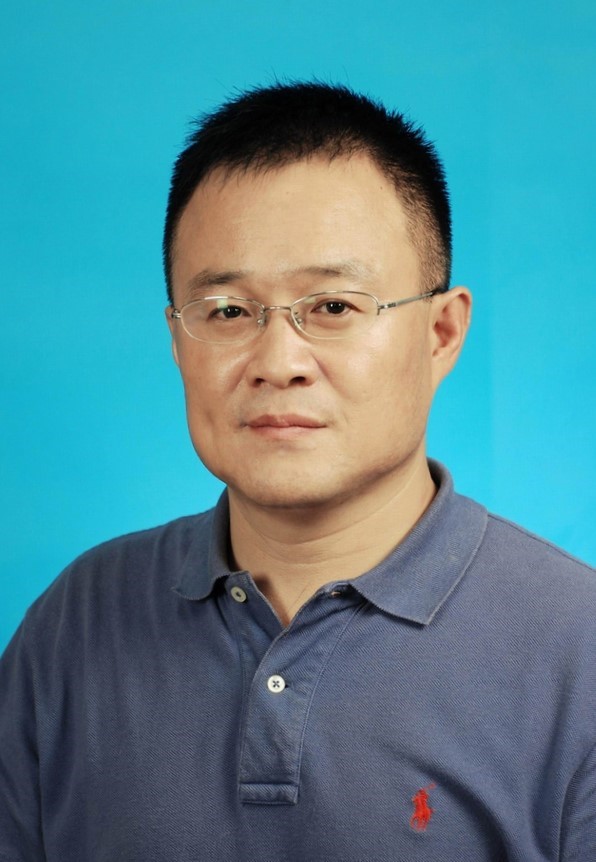}}]
{Yong-Jie Li} Yong-Jie Li (Senior Member, IEEE) received the Ph.D. degree in biomedical engineering from
UESTC, in 2004. He is currently a Professor with the Key Laboratory for NeuroInformation of Ministry of
Education, School of Life Science and Technology, University of Electronic Science and Technology of
China. His research focuses on building of biologically inspired computational models of visual perception and 
the applications in image processing and computer vision.
\end{IEEEbiography}
\vfill

\end{document}